\pdfoutput=1

\documentclass[11pt]{article}
\usepackage[final]{acl}

\usepackage{times}
\usepackage{latexsym}

\usepackage[T1]{fontenc}
\usepackage[utf8]{inputenc}
\usepackage{microtype}
\usepackage{inconsolata}
\usepackage{graphicx}

\usepackage{amsmath}
\usepackage{tabularx}

\usepackage{multirow}
\usepackage{array}

\usepackage{booktabs}
\usepackage{natbib}
\usepackage{float}
\usepackage{longtable}
\usepackage{graphicx}
\usepackage{url}

\newcommand{\ours}{\textsc{NexusSum}}

\title{\ours: Hierarchical LLM Agents for \\ Long-Form Narrative Summarization}

\author{
  Hyuntak Kim\thanks{Equal contribution.}\quad
  Byung-Hak Kim\footnotemark[1]~\thanks{Corresponding author: \texttt{bhak.kim@cj.net}} \\[1ex]
  CJ Corporation \\[0.5ex]
}

\begin{document}

\maketitle

\begin{abstract}

Summarizing long-form narratives—such as books, movies, and TV scripts—requires capturing intricate plotlines, character interactions, and thematic coherence, a task that remains challenging for existing LLMs.
We introduce \ours{}, a multi-agent LLM framework for narrative summarization that processes long-form text through a structured, sequential pipeline—without requiring fine-tuning. Our approach introduces two key innovations:
\textbf{(1) Dialogue-to-Description Transformation}: A narrative-specific preprocessing method that standardizes character dialogue and descriptive text into a unified format, improving coherence.
\textbf{(2) Hierarchical Multi-LLM Summarization}: A structured summarization pipeline that optimizes chunk processing and controls output length for accurate, high-quality summaries.
Our method establishes a new state-of-the-art in narrative summarization, achieving up to \textbf{a 30.0\% improvement in BERTScore (F1)} across books, movies, and TV scripts. These results demonstrate the effectiveness of multi-agent LLMs in handling long-form content, offering a scalable approach for structured summarization in diverse storytelling domains.

\end{abstract}

\section{Introduction}
\label{sec:introduction}

\begin{figure}
    \centering
    \includegraphics[width=0.48\textwidth]{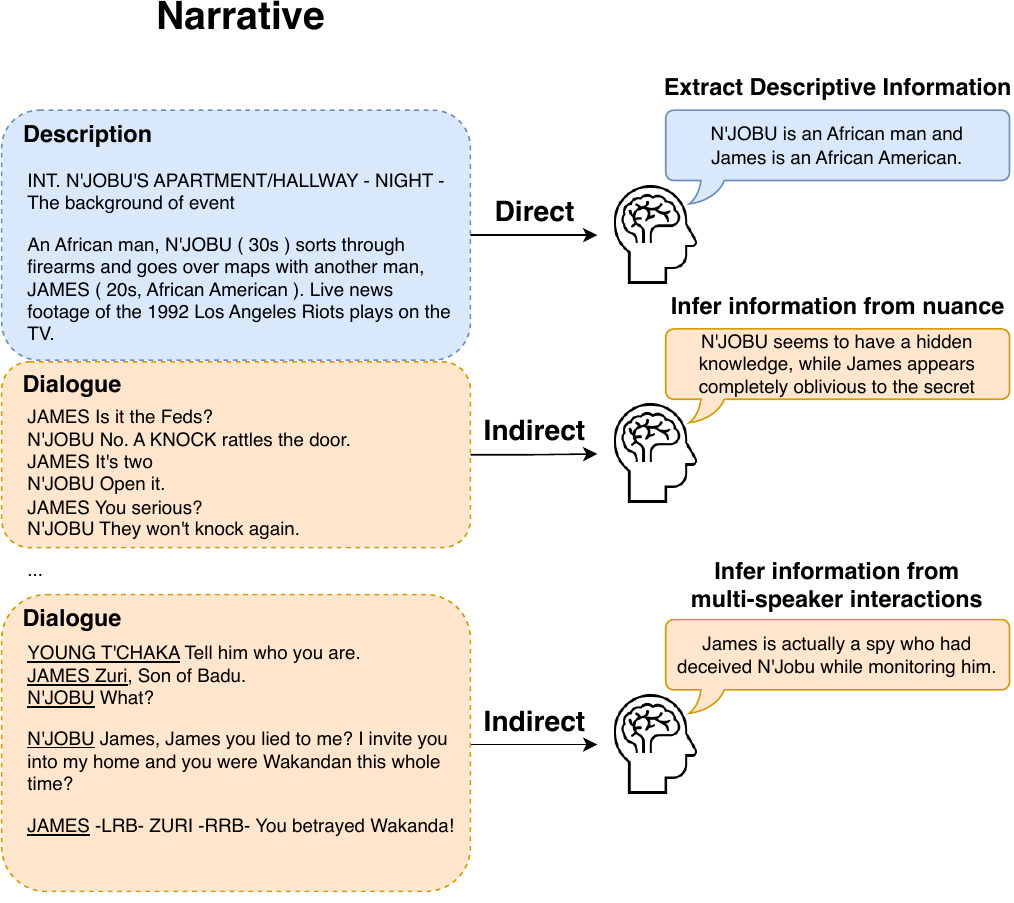}
    \caption{Illustration of narrative structure, showcasing the interplay between descriptive text and multi-speaker dialogues. \ours~enhances coherence by converting dialogues into structured prose for improved long-form summarization.} 
    \label{fig:narrative_feature}
\vspace{-10pt}
\end{figure}

\begin{figure*}
    \centering
    \includegraphics[width=0.95\textwidth]{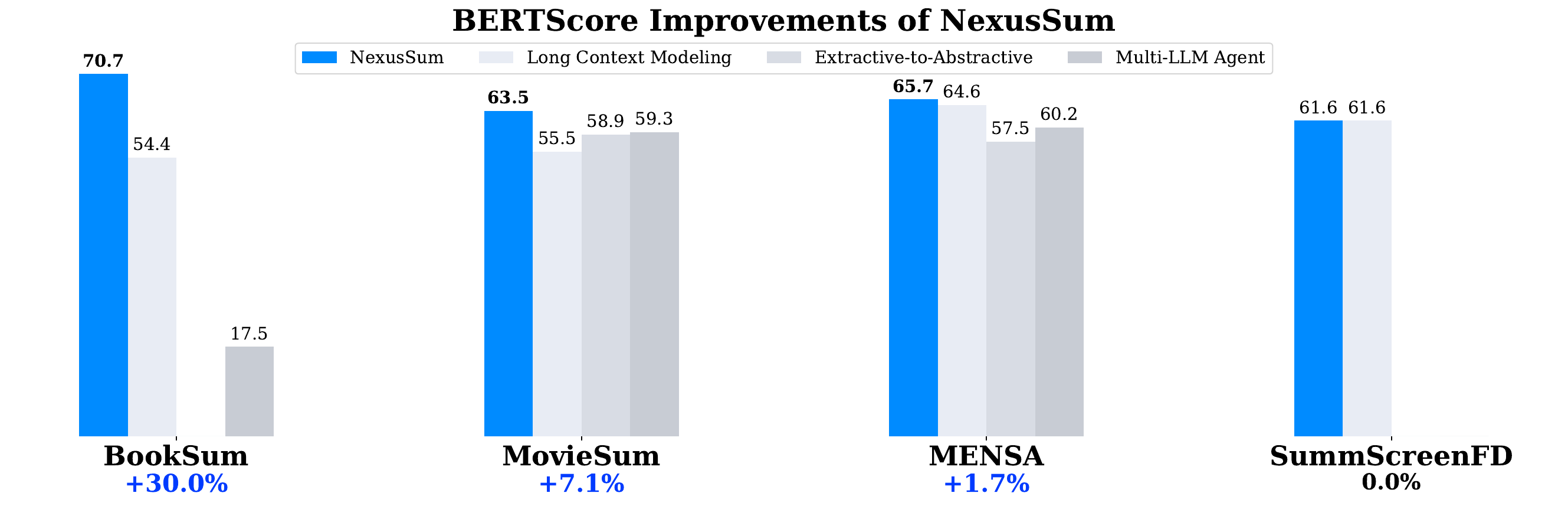}
    \vspace{-5pt}
    \caption{Performance comparison of \ours{} with state-of-the-art summarization models using BERTScore (F1) across multiple benchmarks. \ours{} achieves up to a 30.0\% improvement, particularly excelling in BookSum, where hierarchical processing mitigates context truncation, demonstrating its advantage in long-form narrative summarization. Bold values indicate the new state-of-the-art score.}
    \label{fig:berscore_comparison}
\vspace{-10pt}
\end{figure*}

Summarizing long-form narratives, such as books, movies, and TV scripts, remains an open challenge in NLP. Unlike news or document summarization, narratives require capturing intricate plotlines, evolving character relationships, and thematic coherence over tens of thousands of tokens~\cite{zhao-etal-2022-narrasum}. The hybrid structure of narratives, which combines descriptive prose with multi-speaker dialogues, implicit inference, and dynamic topic shifts (see Figure~\ref{fig:narrative_feature}), adds further complexity~\citep{khalifa-etal-2021-bag, Zou_Zhao_Kang_Lin_Peng_Jiang_Sun_Zhang_Huang_Liu_2021, chen-etal-2022-dialogsum, saxena-keller-2024-moviesum}, demanding an approach that preserves contextual integrity while condensing information effectively. Furthermore, the sheer length of narrative texts, typically ranging from 40K to 160K tokens~\citep{kryscinski-etal-2022-booksum,saxena-keller-2024-select,saxena-keller-2024-moviesum}, poses significant challenges for standard summarization models.

Despite advances in large language models (LLMs) for abstractive summarization \cite{pu2023summarizationalmostdead}, existing methods struggle with long-form narratives for three key reasons. First, context window limitations in LLMs (even with 200K-token capacities~\citep{anthropic2024claude3,gpt4_technical,mistralai2024large}) lead to information loss when processing extended narratives~\cite{liu-etal-2024-lost}. Second, extractive-to-abstractive pipelines~\citep{ladhak-etal-2020-exploring,pu-etal-2022-two,saxena-keller-2024-select} mitigate input constraints by selecting salient sections, but risk omitting critical details, disrupting narrative coherence. Third, zero-shot LLM approaches perform poorly compared to fine-tuned models in narrative summarization~\citep{saxena-keller-2024-select,saxena2025endtoendlongdocumentsummarization}, indicating the need for task-specific adaptations beyond prompt engineering.

Recent Multi-LLM agent frameworks~\citep{guo2024largelanguagemodelbased,zhang2024chain,chang2024booookscore} have introduced strategies to handle long-context documents through segmented inference and hierarchical processing. However, these studies focus primarily on generic document summarization and lack domain-specific optimizations for narrative discourse, character-driven coherence, and length-controlled output generation.

In this work, our aim is to address these challenges by investigating:
\begin{itemize}
    \item \textbf{RQ1} How can a Multi-LLM agent system be designed to summarize long-form narratives while preserving narrative structure and coherence?  \vspace{-1pt} 
    \item \textbf{RQ2} What impact does dialogue-to-descrip-tion transformation have on improving summarization consistency and readability? 
    \item \textbf{RQ3} How does iterative compression affect summary length control and content retention?
\end{itemize} 
To address these challenges, we introduce~\ours, a hierarchical multi-agent LLM framework for long-form narrative summarization.~\ours~employs a three-stage sequential pipeline to progressively refine summaries without fine-tuning:
\begin{itemize}
    \item \textbf{Dialogue-to-Description Transformation} Preprocessor agent converts character dialogues into structured narrative prose, reducing fragmentation and improving coherence.
    \item \textbf{Hierarchical Summarization} Narrative Summarizer agent generates an initial comprehensive summary, preserving key plot points and character interactions.
    \item \textbf{Iterative Compression} Compressor agent dynamically reduces summary length through controlled compression, ensuring key information retention while enforcing length constraints.
\end{itemize}

By segmenting long inputs into manageable chunks and applying hierarchical processing across multiple LLM agents,~\ours~ensures high-fidelity summarization with scalable length control. 
We evaluate~\ours~on four long-form narrative benchmarks: BookSum~\cite{kryscinski-etal-2022-booksum}, MovieSum~\cite{saxena-keller-2024-moviesum}, MENSA~\cite{saxena-keller-2024-select}, and SummScreenFD~\cite{chen-etal-2022-summscreen}. As shown in Figure~\ref{fig:berscore_comparison},~\ours~outperforms existing methods, achieving up to a 30.0\% improvement in BERTScore (F1) \cite{Zhang2020BERTScore:} over previous state-of-the-art models. Our work makes the following contributions:
\begin{itemize}
    \item \textbf{Dialogue-to-Description Transformation} We introduce a novel LLM-based preprocessing step that improves narrative coherence by converting dialogue into structured prose, reducing ambiguity in multi-speaker interactions.
    \item \textbf{Hierarchical Multi-Agent Summarization} We design a structured LLM agent pipeline that refines summaries iteratively, mitigating information loss while preserving contextual dependencies.
    \item \textbf{Optimized Length Control and Chunk Processing} Our framework employs iterative compression and dynamically adjusts chunk sizes, ensuring factual consistency while improving summary conciseness.
    \item \textbf{State-of-the-Art Results} Our approach establishes new benchmarks for long-form narrative summarization, achieving higher accuracy, coherence, and length control than existing LLM-based summarization methods.
\end{itemize}

By advancing multi-LLM agent frameworks for domain-specific narrative summarization,~\ours~provides a scalable, fine-tuning-free solution that enhances long-context understanding across diverse storytelling mediums.

\section{Related Work}
\label{sec:relatedwork}

Narrative summarization differs from traditional document summarization, requiring specialized techniques to handle complex plots, evolving characters, and mixed prose-dialogue structures. This section reviews related work on narrative summarization, long-context summarization, and multi-agent LLMs, positioning~\ours~within this research landscape.

\subsection{Narrative Summarization}
Benchmark datasets like BookSum, MENSA, MovieSum and SummScreenFD have advanced long-form narrative summarization research. Traditional extractive-to-abstractive pipelines~\citep{ladhak-etal-2020-exploring, pu-etal-2022-two} risk losing coherence by omitting character arcs and event dependencies. To address this, scene-based and discourse-aware techniques leverage graph-based models~\cite{gorinski-lapata-2015-movie} and transformer-based saliency classifiers~\cite{saxena-keller-2024-select}. However, these methods struggle with full text processing, often truncating key content. Our approach overcomes this gap by introducing the dialogue-to-description transformation, allowing for a holistic narrative processing while preserving coherence.

\subsection{Long-Context Summarization}
Long-context summarization techniques typically fall into two categories:
\paragraph{\textbf{Architectural Optimization}} Transformer models struggle with scalability due to the quadratic cost of self-attention. Solutions include sparse attention, memory-efficient encoding, and long-context finetuning~\citep{zaheer2020bigbird, Beltagy2020Longformer, kitaev2020reformerefficienttransformer, guo-etal-2022-longt5, wang2020linformerselfattentionlinearcomplexity}. Expanded context windows (up to 200K tokens)~\citep{chen2023extendingcontextwindowlarge, gpt4_technical, mistralai2024large} help but still degrade in multi-turn dependencies, entity tracking, and coherence~\cite{liu-etal-2024-lost}. 
\paragraph{Chunking-Based Method} Chunking-based approaches like SLED~\cite{ivgi-etal-2023-sled} and Unlimiformer~\cite{bertsch2023unlimiformer} segment text for hierarchical summarization, while CachED~\cite{saxena2025endtoendlongdocumentsummarization} improves efficiency via gradient caching but requires finetuning. 

Unlike prior methods, \ours~offers a training-free alternative leveraging Multi-LLM agents, allowing full text summarization without truncation.

\subsection{Multi-Agent LLMs for Summarization}
Recent multi-agent LLM frameworks, such as Chain of Agents (CoA)~\cite{zhang2024chain} and BooookScore~\cite{chang2024booookscore}, improve document summarization through hierarchical merging and sequential refinement (HM-SR) ~\cite{jeong2025agentasjudgefactualsummarizationlong}. However, they lack adaptations for narrative coherence, character interactions, and event dependencies. Retrieval-augmented generation~\cite{NEURIPS2020_rag} improves factuality but struggles with long-form storytelling, often missing thematic continuity~\citep{geng-etal-2022-improving-abstractive, uthus-ni-2023-rise}. \ours~addresses these gaps by integrating the dialogue-to-description transformation and systematic length control, ensuring coherent and contextually faithful summaries.

\section{\ours{} Framework}
\label{sec:relatedworks}

\begin{figure*}
    \centering
    \includegraphics[width=0.85\textwidth]{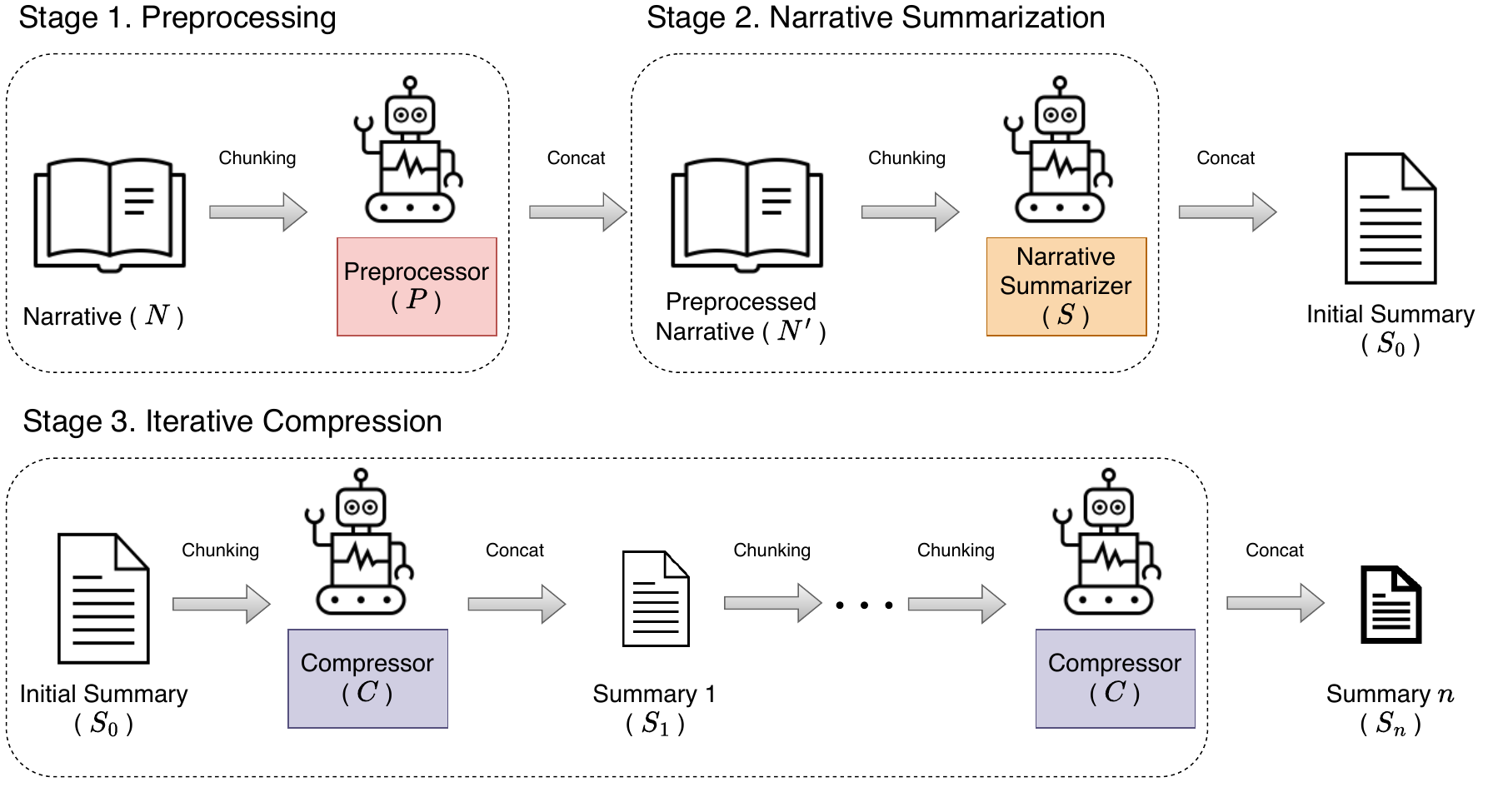}
    \vspace{-10pt}
    \caption{Overview of \ours{}, a hierarchical multi-agent LLM framework for long-form narrative summarization. It follows a three-stage pipeline: (1) \textbf{Preprocessing} converts dialogues into descriptive prose, (2) \textbf{Narrative Summarization} generates an initial summary, and (3) \textbf{Iterative Compression} refines it for length control while preserving key details.}
    \label{fig:overview}
\vspace{-10pt}
\end{figure*}

To address the challenges of long-form narrative summarization, we introduce \ours{}, a hierarchical multi-agent LLM framework that processes narratives through a three-stage pipeline: \textbf{Preprocessing}, \textbf{Narrative Summarization}, and \textbf{Iterative Compression}. The system is designed to preserve narrative coherence, optimize summary length, and ensure information retention without requiring fine-tuning. Figure~\ref{fig:overview} provides a schematic of the framework. Each stage of~\ours~is optimized using a chunk-and-concat method, allowing for scalable summarization of narratives of arbitrary length while ensuring controlled compression. We detail the functionality of each stage below.\footnote{For a comprehensive breakdown of each stage, see Appendix~\ref{appendix:prompt} for the prompts used by each agent, and Appendix~\ref{appendix:intermediate_samples} for illustrative output samples.}

\subsection{Preprocessing Stage}
Narrative texts combine dialogues and descriptions, often leading to fragmented summaries. The \emph{Preprocessor} agent \(P\) enhances coherence by converting dialogues into structured third-person prose, simplifying input for summarization. Following~\cite{xu-etal-2022-narrate}, we prompt an LLM to reframe dialogues while preserving the intent of the speaker.

To manage long input lengths efficiently, \(P\) segments the input text into scene-based chunks, following recent studies~\citep{saxena-keller-2024-select,jeong2025agentasjudgefactualsummarizationlong} that demonstrate the effectiveness scenes as semantic units for processing narratives:
\begin{equation} 
    N = n_1 \oplus n_2 \oplus \cdots \oplus n_k 
\end{equation}
where \(N\) represents the input narrative, segmented into \(k\) chunks. The number of chunks \(k\) is dynamically computed based on a fixed scene-based chunk size\footnote{We set the chunk size to 8 scenes, balancing context retention and processing efficiency, resulting in \(k = \text{total scenes} / 8\).}, and \(\oplus\) denotes concatenation. Once processed, the output is a preformatted narrative text \(N'\), ready for summarization:
\begin{equation}
    N' = P(n_1) \oplus P(n_2) \oplus \cdots \oplus P(n_k).
\end{equation}

\subsection{Narrative Summarization}
The \emph{Narrative Summarizer} agent \(S\) generates an initial abstract summary from the preprocessed text \(N'\). To maintain coherence across long documents, \(N'\) is further chunked into scene-based units:
\begin{equation}
    N' = n'_1 \oplus n'_2 \oplus \cdots \oplus n'_j
\end{equation}
where \(j\) is the number of chunks\footnote{For simplicity, we set \( j = k \), ensuring each chunk contains eight scenes.}. The summarization process follows:
\begin{equation}
    S_0 = S(n'_1) \oplus S(n'_2) \oplus \cdots \oplus S(n'_j)
\end{equation}
where \(S_0\) represents the initial summary. Unlike traditional single-pass models, \ours{} applies hierarchical chunk processing, allowing long-range information retention.

\begin{table*}[t] 
    \centering 
    \small 
    \setlength{\tabcolsep}{4pt} 
    \renewcommand{\arraystretch}{0.80} 
    \begin{tabular}{l|cccc} 
        \toprule 
        \textbf{Dataset} & \textbf{BookSum} & \textbf{MovieSum} & \textbf{MENSA} & \textbf{SummScreenFD} \\ 
        \midrule 
        Domain & Novels & Movies & Movies & TV Shows \\ 
        Eval Dataset Count & 17 & 200 & 50 & 337 \\ 
        \midrule 
        Avg. Input Length (Tokens) & 158,645 (98.06\%) & 42,999 (24.08\%) & 39,808 (21.27\%) & 9,464 (38.91\%) \\ 
        Avg. Output Length (Tokens) & 1,792 (46.43\%) & 902 (26.05\%) & 952 (17.02\%) & 151 \textbf{(76.16\%)} \\ 
        \bottomrule 
    \end{tabular} 
    \caption{Overview of four narrative summarization datasets, highlighting diverse text structures and summary styles. Input and output lengths are reported with Coefficient of Variation (CV), with SummScreenFD’s high CV (76.16\%) indicating significant variability, making it a challenging benchmark for consistency.} 
    \vspace{-10pt}
    \label{tab:dataset_stats} 
\end{table*}
\subsection{Iterative Compression} 
While \(S_0\) is an informative summary, it may exceed the desired length constraints. The \emph{Compressor} agent \(C\) applies iterative compression to refine \(S_0\) while preserving key narrative details. Our iterative compression method consists of two steps: sentence-based chunking followed by hierarchical compression.

\paragraph{Sentence-Based Chunking}
Unlike the previous scene-based chunking, compression requires sentence-level granularity. We divide \(S_0\) into smaller units:
\begin{equation}
    S_0 = s_{0,1} \oplus s_{0,2} \oplus \cdots \oplus s_{0,l_{0}}
\end{equation}
where \(l_0\) is the number of chunks in the initial text, which is dynamically adjusted to maintain optimal compression ratios. Sentences are grouped into chunks up to a predetermined token size \(\delta\), allowing flexible compression rates. This \(\delta\) plays a crucial role in controlling the compression ratio of our system's output, as the smaller size of input yields lower compression (see our empirical analysis in Appendix~\ref{appendix:impact_of_chunk_size}).

\paragraph{Hierarchical Compression}
Following chunking, we apply hierarchical compression iteratively. In each iteration \(i\), the Compressor agent \(C_i\) refines the previous compressed summary \(S_{i-1}\), which is split into \(l_{i-1}\) chunks by Sentence-Based Chunking. The \(i\)-th Compressor agent \(C_i\) iteratively refines the summary:
\begin{equation}
    \scalebox{0.9}{$S_i = C_i(s_{i-1,1}) \oplus C_i(s_{i-1,2}) \oplus \cdots \oplus C_i(s_{i-1,l_{i-1}})$}.
\end{equation}

The process continues for \(n\) iterations, dynamically determined by a target word count \(\theta\)\footnote{If the output is already below \(\theta\), the final summary may also be shorter than the target.}: 
\begin{itemize}
    \item If \(S_i\) exceeds \(\theta\), compression continues.
    \item If \(S_i\) falls below \(\theta\), the previous iteration’s output is used.
\end{itemize}
To balance quality and computational efficiency, we limit compression to a maximum of 10 iterations.

\section{Experimental Setup}
\label{sec:experimentalsetup}
This section describes four datasets of narrative summarization benchmarks, state-of-the-art baselines, implementation details, and evaluation metrics used in our study to evaluate \ours{}.
\subsection{Dataset}
We conducted experiments on four diverse long-form narrative summarization datasets covering novels, movies, and TV scripts. Table~\ref{tab:dataset_stats} summarizes the key statistics of the dataset.

\paragraph{Dataset Descriptions} 
\begin{itemize}
    \item BookSum: A novel-based summarization dataset with the longest input and output sequences requiring strong long-context comprehension.
    \item MovieSum: Contains summaries of 200 movies with moderate-length documents.
    \item MENSA: A script-based dataset combining ScriptBase~\cite{gorinski-lapata-2015-movie} and recent movie scripts, providing rich character interactions and scene-based storytelling.
    \item SummScreenFD: A dataset from TV shows with concise and highly variable summaries (CV\footnote{CV measures dispersion calculated as the ratio of the standard deviation to the mean, expressed as a percentage.} = 76.16\%), testing the adaptability of \ours{} to various writing styles.
\end{itemize}

\subsection{Baselines}
We compare \ours{} with three main baseline categories, covering long context modeling, extractive-to-abstractive methods, and Multi-LLM agent frameworks.

\paragraph{Long Context Modeling Baselines} These approaches modify model architectures to handle extended sequences:
\begin{itemize}
    \item Zero-Shot through GPT-4o~\cite{gpt4_technical} and Mistral-Large~\cite{mistralai2024large}: Uses maximum context window expansion but struggles with truncation.
    \item SLED~\cite{ivgi-etal-2023-sled}: Uses local attention with a sliding window mechanism.
    \item Unlimiformer~\cite{bertsch2023unlimiformer}: Extends transformers with unlimited retrieval-based attention.
    \item CachED~\cite{saxena2025endtoendlongdocumentsummarization}: A gradient caching approach for memory-efficient summarization.
\end{itemize}

\begin{table*}[t]
\centering
\footnotesize
\setlength{\tabcolsep}{2pt} 
\renewcommand{\arraystretch}{0.90} 
\begin{tabular}{lcccc}
\toprule
\textbf{Method} & \textbf{BookSum} & \textbf{MovieSum} & \textbf{MENSA} & \textbf{SummScreenFD} \\
\midrule
\multicolumn{5}{l}{\textbf{Long Context Modeling}} \\
Zero-Shot (Mistral Large, 123B) & 46.42 & 55.50 & 54.80 & 57.23 \\
Zero-Shot (GPT4o) & 47.24 & - & 52.8 & - \\
SLED (BART Large, 406M) & 52.4 & - & 58.3 & 59.9 \\
Unlimiformer (BART Base, 139M) & 51.5 & - & 58.7 & 58.5 \\
CachED (BART Large, 406M) & \underline{54.4} & - & \underline{64.6} & \underline{\textbf{61.59}} \\
\midrule
\multicolumn{5}{l}{\textbf{Extractive-to-Abstractive}} \\
Description Only (LED-Large, 459M) & - & 58.92 & - & - \\
Two-Stage Heuristic (LED-Large, 459M) & - & 58.54 & 56.34 & - \\
Summ N (LED-Large, 459M) & - & - & 40.87 & - \\
Select and Summ (LED-Large, 459M) & - & - & 57.46 & - \\
\midrule
\multicolumn{5}{l}{\textbf{Multi-LLM Agent}} \\
HM-SR (GPT4o-mini) & - & \underline{59.32} & 60.22 & - \\
CoA (Claude 3 Opus) & \underline{(17.47)} & - & - & - \\
\midrule
\textbf{\ours} (Mistral Large, 123B) 
& \textbf{(18.27)} / \textbf{70.70} 
& \textbf{63.53} 
& \textbf{65.73} 
& \textbf{61.59*} \\
\bottomrule
\end{tabular}
\vspace{-5pt}
\caption{Performance comparison of \ours{} with state-of-the-art summarization models using BERTScore (F1) and ROUGE (geometric mean of 1/2/L) in parentheses. \ours{} achieves its highest gains on BookSum (+30.0\%) and MovieSum (+7.1\%), outperforming Multi-LLM baselines like CoA and HM-SR. Baseline results are sourced from previous studies~\citep{saxena-keller-2024-moviesum,saxena-keller-2024-select,saxena2025endtoendlongdocumentsummarization,jeong2025agentasjudgefactualsummarizationlong,zhang2024chain}. For open-source models, parameter sizes are shown in parentheses as (Model, Size).}
\vspace{-15pt}
\label{tab:results}
\end{table*}

\paragraph{Extractive-to-Abstractive Baselines} These approaches extract salient segments before abstractive summarization:
\begin{itemize}
    \item Description Only~\cite{saxena-keller-2024-moviesum}: Selects descriptive sections for summarization.
    \item Two-Stage Heuristics~\cite{pu-etal-2022-two}: Extracts character actions and key dialogues.
    \item Summ N~\cite{zhang-etal-2022-summn}: Generates coarse summaries, then refines outputs iteratively.
    \item Select and Summ~\cite{saxena-keller-2024-select}: Uses scene saliency classifiers to extract important moments.
\end{itemize}

\paragraph{Multi-LLM Agent Frameworks} 
\begin{itemize}
    \item HM-SR~\cite{jeong2025agentasjudgefactualsummarizationlong}: Applies hierarchical chunk merging with refinement agents.
    \item CoA~\cite{zhang2024chain}: A multi-agent LLM pipeline, where each agent specializes in refining a specific summary aspect.
\end{itemize}

\subsection{Implementation Details}
We implement \ours{} using Mistral-Large-Instruct-2407 (123B)~\cite{mistralai2024large}, with optimized inference via vLLM~\cite{kwon2023efficient} with temperature = 0.3, top-p = 1.0 and seed = 42. The model is run on four A100 GPUs. For Claude 3 Haiku~\cite{anthropic2024claude3}, we set temperature = 0 to minimize randomness. For each benchmark, \ours{}'s configuration of \(\delta\) and \(\theta\) are detailed in the Appendices~\ref{appendix:optimal_chunk_size} and ~\ref{appendix:optimal_lower_bound}. 

To ensure that our LLM models were not exposed to evaluation datasets during training, we conducted an n-gram overlap analysis (see Appendix~\ref{appendix:dataset_contamination}). The results confirmed that the overlap remained below 2\% on all benchmarks, indicating minimal data leakage and an unbiased evaluation.

\subsection{Evaluation Metrics}
We evaluate ~\ours{} using a semantic similarity, length control metrics.
\begin{itemize}
    \item \textbf{BERTScore (F1)} measures semantic similarity beyond n-gram overlap, aligning with human judgement. We use DeBERTa-XLarge-MNLI~\cite{he2021deberta} as the base model, following established practices~\citep{saxena2025endtoendlongdocumentsummarization, saxena-keller-2024-select}. ROUGE (1/2/L) scores~\cite{lin-2004-rouge} are reported in Appendix~\ref{appendix:rouge_scores} for comparability with prior work, though BERTScore better captures abstraction quality.
    \item \textbf{Length Adherence Rate (LAR)} measures the degree to which a summary matches the target word counts, defined as
    \begin{equation}
        \text{LAR} = 1 - |L_{\text{gen}} - L_{\text{target}}| \times L_{\text{target}}^{-1}
    \end{equation}
    to quantify the effectiveness of iterative compression in controlling summary length.
\end{itemize}

\section{Results and Analysis}
\label{sec:resultsandanalysis}
We evaluate \ours{} against state-of-the-art baselines on four narrative summarization benchmarks, assessing performance gains, ablation results, length control and agent adaptability. Additional analyses on factuality, document utilization and inference time complexity are provided in Appendices~\ref{appendix:factuality_performance},~\ref{appendix:document_utilization} and~\ref{appendix:time_complexity}.

\subsection{Benchmark Performance}
Table~\ref{tab:results} summarizes the results, showing that \ours~outperforms all baselines across datasets, achieving state-of-the-art performance with  substantial improvements over prior methods:

\paragraph{BookSum} \ours~outperforms CachED by +30.0\% BERTScore (F1), showcasing its effectiveness in processing extended narratives without context loss. Unlike CachED's static chunking approach, \ours~dynamically optimizes chunk sizes and applies iterative compression, preserving key information while enhancing coherence in long-form summarization. Additionally, \ours~surpasses CoA by +4.6\% in ROUGE (geometric mean of 1/2/L), despite CoA leveraging Claude-3-Opus~\cite{anthropic2024claude3}, a top-performing model for long-context summarization.

\paragraph{MovieSum} \ours~outperforms HM-SR by +7.1\%, leveraging structured length control to maintain consistency in multi-scene script summaries. While HM-SR effectively merges hierarchical summaries, its lack of precise length constraints results in variable-length outputs.

\paragraph{MENSA} \ours~achieves a +1.7\% gain over CachED, surpassing long-context models in screenplay summarization. It also outperforms Select and Summ by +14.4\%, demonstrating superior abstraction for character-driven plots, where even extractive-to-abstractive methods struggle with maintaining narrative depth beyond scene selection.

\paragraph{SummScreenFD} \ours~matches the performance of CachED while ensuring better length control through iterative compression, reducing output variability compared to zero-shot baselines.

\begin{table}[t]
\small 
\setlength{\tabcolsep}{2pt} 
\renewcommand{\arraystretch}{0.90} 
\centering
\begin{tabular}{l|c|c}
\toprule
\textbf{Method} & \textbf{BERTScore (F1)} & \textbf{Improvement} \\
\midrule
Zero-Shot & 54.81 & - \\
\(P\) + Zero-Shot & 57.26 & +2.45 \\
\(P\) + \(S\) & 62.12 & +4.86 \\
\(S\) + \(C\) & 63.90 & +1.78 \\
\(P\) + \(S\) + \(C\) (\ours) & \textbf{65.73} & +1.83 \\
\bottomrule
\end{tabular}
\vspace{-5pt}
\caption{Ablation analysis on the MENSA dataset, showing contributions of each LLM agent stage to a final BERTScore (F1) of 65.73.}
\vspace{-15pt}
\label{tab:ablation}
\end{table}

\begin{table}[t]
\centering
\footnotesize
\setlength{\tabcolsep}{4pt} 
\renewcommand{\arraystretch}{0.70} 
\begin{tabular}{l|cccc}
\toprule
{\textbf{Target Length}} & \textbf{600} & \textbf{900} & \textbf{1200} & \textbf{1500} \\
\midrule
\multicolumn{5}{l}{\textbf{Length (Word)}} \\
Zero-Shot & 453 & 540 & 571 & 592 \\
Ours & \textbf{670} & \textbf{891} & \textbf{1385} & \textbf{1621} \\
\midrule
\multicolumn{5}{l}{\textbf{LAR}} \\
Zero-Shot & 0.245 & 0.400 & 0.524 & 0.605 \\
Ours & \textbf{0.883} & \textbf{0.990} & \textbf{0.988} & \textbf{0.914} \\
\midrule
\multicolumn{5}{l}{\textbf{BERTScore (F1)}} \\
Zero-Shot & 56.85 & 58.18 & 58.55 & 57.75 \\
Ours & \textbf{63.59} & \textbf{65.73} & \textbf{65.21} & \textbf{62.86} \\
\bottomrule
\end{tabular}
\vspace{-5pt}
\caption{
Comparison of \ours{} and Zero-Shot summarization on length control, measured by word count deviation, LAR and BERTScore (F1). \ours{} achieves a higher BERTScore while maintaining an LAR close to 1.0, demonstrating precise adherence to target length constraints.}
\vspace{-5pt}
\label{tab:length_control}
\end{table}

\label{subsec:adaptive_performance}

\begin{table}[t]
\centering
\small 
\setlength{\tabcolsep}{4pt} 
\renewcommand{\arraystretch}{0.90} 
\begin{tabular}{l|c}
\toprule
\textbf{Method} & \textbf{BERTScore (F1)} \\
\midrule
\ours~\textsubscript{base} & 56.61 \\
\ours~\textsubscript{CoT} & 58.61 \\
\ours~\textsubscript{CoT+FewShot} & \textbf{61.59} \\
\bottomrule
\end{tabular}
\vspace{-5pt}
\caption{Effect of prompt engineering on \ours{} performance in SummScreenFD. Incorporating CoT and Few-Shot learning results in a 5.0-point BERTScore improvement, highlights \ours{}'s adaptability to diverse summarization styles without parameter updates.}
\vspace{-10pt}
\label{tab:adaptation_results}
\end{table}

\subsection{Contribution of LLM Agents}
\label{subsec:ablation_analysis_of_llm_agent_stages}

Table~\ref{tab:ablation} quantifies the contribution of each \ours{} component through an ablation study. Zero-Shot summarization serves as the baseline, achieving BERTScore of 54.81. Introducing \(P\) improves coherence by converting dialogues into narrative text, raising BERTScore to 57.26 (+2.45). This confirms that \(P\) is insufficient alone for high-quality summarization. The addition of \(S\) further improves BERTScore to 62.12 (+4.86). Finally, \(C\) refines summary length while retaining critical details, producing the highest performance of 65.73. These results validate that each component of \ours{} contributes to performance improvements, with the multi-agent LLM framework being essential for long-form narrative coherence and retention.

\subsection{Length Control with Quality Preservation}
\label{subsec:length_control_with_quality_preservation}

We evaluated \ours{}'s length control capabilities on the MENSA dataset. As a baseline, we use a Zero-Shot model with explicit length constraints applied via prompt instructions (\emph{"Write in [target length]"}). As shown in Table~\ref{tab:length_control}, \ours{} effectively balances semantic quality (BERTScore (F1)) and length adherence (LAR) in all target lengths. This shows that \ours{} not only generates more semantically accurate summaries, but also enforces structured length control effectively than conventional prompting strategies.

\subsection{Adaptive Performance through Prompt Engineering}

As a Multi-LLM agent framework, \ours{} leverages prompt engineering to adapt to diverse summarization tasks without requiring parameter updates. We evaluate this adaptability on SummScreenFD, a challenging dataset characterized by spoken dialogue format and high variable summary styles (CV= 76.16\%, see Table~\ref{tab:dataset_stats}). To enhance adaptation, we incorporate Chain of Thoughts (CoT) reasoning~\cite{kojima2023largelanguagemodelszeroshot} in \(P\) and Few-Shot learning~\cite{few-shot} in \(S\) and \(C\) to refine output style.

Table \ref{tab:adaptation_results} demonstrates that CoT alone improves BERTScore (F1) from 56.61 to 58.61 (+2.0 points), while adding Few-Shot learning further boosts performance to 61.59 (+2.98 points). These results highlight \ours{}'s ability to adapt to diverse summarization scenarios using simple prompt customization, ensuring robust generalization across narrative structures without additional training or fine-tuning.

\subsection{Human Preference Analysis}
\label{subsec:human_preference}
\paragraph{Setup} To explore human preference for generated summaries across different narrative styles and genres, we create three different K-Drama summaries that vary in genres (Fantasy-Romance, Korean History and Modern Romantic-Comedy). Summaries were generated using three different methodologies (Zero-Shot, \ours~ and \ours\textsubscript{R}) to enable a comparative preference analysis.

A total of three K-Drama experts participate in the evaluation, with at least two evaluators assessing each output for a given work. They score the summaries on a 5-point Likert scale (1 = Not at all, 5 = Very much so) across four criteria:
\begin{itemize}
    \item \textbf{Key Events}: Are the key events included? 
    \item \textbf{Flow}: Is the contextual information demonstrated specifically?
    \item \textbf{Factuality}: Does the summary have high factual accuracy?
    \item \textbf{Readability}: Does the summary have high readability?
\end{itemize}
In addition, all three evaluators provided qualitative comments to explain the reasons behind their scores (See Appendix~\ref{appendix:qualitative_analysis}).

\paragraph{Results} First, we compare the Zero-Shot method with \ours. Each method aims to generate summaries with a target length of 600 words. Zero-Shot is prompted to generate summaries with the target length of 600 as specified instruction in the prompt. \ours~ generates summaries by halting the iteration process at a lower bound \(\theta=600\).  As shown in Table~\ref{tab:expert_evaluation_results}, \ours~ demonstrates superior summary length control, achieving an average summary length of 609 words, compared to Zero-Shot, which produces summaries with an average length of 219 words. \ours~outperforms Zero-Shot in capturing key events (4.17), maintaining narrative flow (3.34), and ensuring factual accuracy (4). However, Zero-Shot demonstrates superior readability (4.17).

To further enhance readability, we introduce a third method, \ours\textsubscript{R}. This approach incorporates an additional LLM agent that rewrites the original \ours~summary to emulate the concise and fluent style characteristic of the Zero-Shot method. The refining agent smooths sentence transitions, adjusts verbosity, and enhances fluency while preserving key narrative details. In Table~\ref{tab:expert_evaluation_results}, \ours\textsubscript{R} improves readability by +1.5 points compared to \ours, bridging the gap between structured factual summarization and human-preferred fluency.

\section{Conclusion}
\label{sec:conclusionfuturework}

\begin{table}[t]
\centering
\small 
\setlength{\tabcolsep}{2pt} 
\renewcommand{\arraystretch}{0.90} 
\begin{tabular}{l|c|c|c}
\toprule
 & \textbf{Zero-Shot} & \textbf{\ours{}} & \textbf{\ours\textsubscript{R}}  \\
\midrule
Key Events & 3.5 & \textbf{4.17} & \textbf{4.17} \\
Flow & 2.83 & 3.34 & 3\\
Factuality & 3.5 & 4 & 3.67\\
Readability & \textbf{4.17} & 2.17 & \underline{3.67} \\
\midrule
Avg. Output Len & 219 & 609 & 234 \\
\bottomrule
\end{tabular}
\caption{
Expert evaluation of Zero-Shot, \ours, and \ours\textsubscript{R} on K-Drama summaries using a target length (\(\theta\)) of 600 words. Scores reflect performance across four criteria. \ours\textsubscript{R} introduces a reflection step to enhance readability while maintaining high content retention.}
\vspace{-10pt}
\label{tab:expert_evaluation_results}
\end{table}

We introduce \ours, a hierarchical multi-agent LLM framework that advances long-form narrative summarization by improving coherence (RQ2), long-context processing (RQ1), and length control (RQ3). Our results demonstrate that structured multi-agent collaboration enhances information retention while maintaining coherence, laying the groundwork for scalable, adaptive AI summarization systems.

Beyond state-of-the-art performance, \ours~has broader implications for AI-driven storytelling, personalized summarization, and conversational AI. Our findings on Chain-of-Thought-driven self-planning suggest a path toward autonomous, context-aware LLM agents capable of refinement without retraining. However, human evaluation highlights a readability gap compared to Zero-Shot baselines. Future work should explore a fluency-enhancing summarization framework while preserving factual consistency and optimizing multi-agent collaboration efficiency.

\section{Limitations}
\label{sec:limitations}

While \ours~introduces significant advancements in long-form narrative summarization, certain limitations remain, particularly in evaluation paradigms, readability, and adaptability. This section outlines key challenges and directions for future improvements.

\paragraph{Limitation of Automated Metrics} 
Automated evaluation metrics such as BERTScore and ROUGE provide useful approximations of summary quality but fail to capture readability, coherence, and user preference, which are critical for long-form narrative summarization. To address these gaps, we conducted an expert evaluation on three K-drama summaries from distinct genres (historical, fantasy, and slice-of-life) to assess readability, coherence, and factual accuracy (Section~\ref{subsec:human_preference}).

As shown in Table~\ref{tab:expert_evaluation_results}, \ours{} produces summaries closer to the target length (609 words) than Zero-Shot (219 words). However, despite \ours{} achieving higher BERTScore and ROUGE, experts rated Zero-Shot outputs as more readable (4.17 vs. 2.17). This discrepancy suggests Zero-Shot favors fluency and stylistic variation at the cost of factual accuracy, whereas \ours{} focuses on key event retention, leading to denser summaries that may feel less natural to human readers. These findings highlight a crucial limitation of current summarization evaluation paradigms—higher automated scores do not necessarily align with human preference.

\paragraph{Future Directions} 
Human feedback (Section~\ref{subsec:human_preference}, Appendix~\ref{appendix:qualitative_analysis}) suggests that \ours\textsubscript{R} reduces rigid phrasing and improves narrative flow, making summaries more natural while retaining essential content. This demonstrates that an additional reflection step can significantly enhance human preference alignment, opening the door to adaptive post-processing techniques for long-form summarization to offer customizable and more engaging user experiences.

\section*{Acknowledgments}
We thank our colleagues at the AI R\&D Division for their insightful discussions that helped shape the direction of this work. We also thank the team at CJ ENM for their support with human evaluation and for providing valuable feedback throughout the project.


\bibliography{acl_2025}



\newpage

\appendix
\onecolumn

\newpage
\section{LLM Agent Prompts in \ours}
\label{appendix:prompt}

\subsection{Preprocessing Stage}
\subsubsection{Preprocessor Agent \texorpdfstring{(\(P\))}{(P)}}

\begingroup 
\small 
\setlength{\tabcolsep}{4pt}
\begin{longtable}{p{0.15\linewidth}p{0.8\linewidth}}
\toprule
\textit{Single-Turn} & \\
\midrule
\textbf{System Prompt} & 
You are an expert script-to-narrative converter. Transform input script into engaging prose narratives while preserving the essence of the original work. \\
\midrule
\textbf{User Prompt} & 
\#\# INPUT\_SCRIPT \newline
[Put your input narrative here]\newline

\#\# Guidelines: \newline
1. Convert dialogue to reported speech, including emotions and speaker traits. \newline
2. Integrate narration and stage directions seamlessly. \newline
3. Preserve original structure, pacing, and character voices. \newline
4. Capture emotional tone and subtext. \newline
5. Use varied language for different speaking styles. \newline
6. Include relevant context from stage directions. \newline
7. Create a cohesive narrative retaining key dramatic elements. \newline
8. Maintain original language and formal written style. \newline
9. Use third-person perspective. \newline

\#\# Output Format: \newline
[Scene Heading: line starting with "INT." or "EXT."] (skip this if there is no scene heading) \newline
[Your Narrative Here] \newline

[Scene Heading: line starting with "INT." or "EXT."] (skip this if there is no scene heading) \newline
[Your Narrative Here] \\
\bottomrule
\caption{Preprocessor Agent Prompt}
\label{tab:preprocessor_agent}
\end{longtable}
\endgroup


\newpage
\subsubsection{Preprocessor Agent with CoT \texorpdfstring{(\(P'\))}{(P')}}

\begingroup 
\small 
\setlength{\tabcolsep}{4pt}
\begin{longtable}{p{0.15\linewidth}p{0.8\linewidth}}
\toprule
\textit{Multi-Turn} & \\
\midrule
\textbf{System Prompt} & 
You are an expert script analyst tasked with creating a tailored strategy to transform a specific script's dialogue into narrative form while preserving essential elements. Your strategy should be based on the unique characteristics of the input script provided. \\
\midrule
\textbf{User Prompt} & 
\#\# INPUT\_SCRIPT \newline
[Put your input narrative here]\newline

\#\# Guidelines: \newline
1. Analyze the provided script carefully, noting its structure, style, and unique features. \newline
2. Identify the types of dialogue present (e.g., conversations, monologues, voice-overs). \newline
3. Recognize the script's format for scene headings, time indicators, and descriptive elements. \newline
4. Determine the overall tone and atmosphere of the script. \newline
5. Create a detailed, step-by-step strategy for transforming this specific script, addressing: \newline
- How to handle the particular dialogue styles present \newline
- Methods to preserve the script's unique formatting and structural elements \newline
- Techniques for maintaining the script's tone and atmosphere in narrative form \newline
- Approaches to seamlessly integrate dialogue with descriptive text \newline

\#\# Output Format: \newline
[Strategy Here] \\
\midrule
\textbf{System Prompt} & 
You are a skilled scriptwriter and narrative expert. Using the strategy provided below, transform the given script by converting dialogue into narrative form while seamlessly integrating it with the original descriptive text. Ensure that you preserve all essential elements such as scene headings, time, background, and event descriptions. \\
\midrule
\textbf{User Prompt} & 
\#\# Strategy: \newline
[Put your strategy here] \newline

\#\# INPUT\_SCRIPT: \newline
[Put your input narrative here]\newline

\#\# Instructions: \newline
1. Transform the dialogue into narrative form, integrating it smoothly with the descriptive text. \newline
2. Follow each step of the provided strategy meticulously. \newline

\#\# Output Format: \newline
[Scene Heading Here] \newline
[Your Narrative Paragraph Here] \newline

[Scene Heading Here] \newline
[Your Narrative Paragraph Here] \newline

... \\ 
\bottomrule
\caption{Preprocessor Agent Prompt with CoT}
\label{tab:preprocessor_agent_cot}
\end{longtable}
\endgroup


\newpage
\subsection{Narrative Summarization Stage}

\subsubsection{Narrative Summarizer Agent \texorpdfstring{(\(S\))}{(S)}}

\begingroup 
\small 
\setlength{\tabcolsep}{4pt}
\begin{longtable}{p{0.15\linewidth}p{0.8\linewidth}}
\toprule
\textit{Single-Turn} & \\
\midrule
\textbf{System Prompt} & 
You are an expert storyteller. Create a concise summary of the given part of the script. \\
\midrule
\textbf{User Prompt} & 
\#\# PART\_OF\_SCRIPT \newline
[Put your input preprocessed narrative here] \newline

\#\# Guidelines: \newline
- Write a summary of the PART\_OF\_SCRIPT. \newline
- Focus on key events, character traits, and interactions. \newline

\#\# Output Format: \newline

\# Summary \newline
[Your summary here] \\
\bottomrule
\caption{Narrative Summarizer Agent Prompt}
\label{tab:narrative_summarizer_agent}
\end{longtable}
\endgroup


\subsubsection{Narrative Summarizer Agent with Few-Shots \texorpdfstring{(\(S'\))}{(S')}}

\begingroup 
\small 
\setlength{\tabcolsep}{4pt}
\begin{longtable}{p{0.15\linewidth}p{0.8\linewidth}}
\toprule
\textit{Single-Turn} & \\
\midrule
\textbf{System Prompt} & 
You are an expert storyteller. Create a concise summary of the given part of the script. Refer to the Example Output to generate the output following their styles. \\
\midrule
\textbf{User Prompt} & 
\#\# PART\_OF\_SCRIPT \newline
[Put your input preprocessed narrative here] \newline

\#\# Examples of Output: \newline

\#\#\# Example Output 01 \newline
[Put your output example] \newline

\#\#\# Example Output 02 \newline
[Put your output example] \newline

\#\#\# Example Output 03 \newline
[Put your output example] \newline

----------------------------- \newline
Now, write a summary of the following PART\_OF\_SCRIPT. \newline

\#\# Guidelines: \newline
- Write a comprehensive summary of the PART\_OF\_SCRIPT. \newline
- Focus on key plot points, character interactions, and significant events. \newline
- Include relevant character names and locations. \newline
- Highlight any central conflicts or challenges faced by the characters. \newline
- Mention any overarching themes or tones (e.g., suspense, comedy, drama). \newline
- Aim for a length of 3-5 sentences. \newline

\#\# Output Format: \newline

\# Summary \newline
[Your summary here] \\
\bottomrule
\caption{Narrative Summarizer Agent Prompt with Few-Shots}
\label{tab:narrative_summarizer_agent_few_shot}
\end{longtable}
\endgroup

\newpage
\subsection{Iterative Compression Stage}

\subsubsection{Iterative Compressor Agent \texorpdfstring{(\(C\))}{((C)}}

\begingroup 
\small 
\setlength{\tabcolsep}{4pt}
\begin{longtable}{p{0.15\linewidth}p{0.8\linewidth}}
\toprule
\textit{Single-Turn} & \\
\midrule
\textbf{System Prompt} & 
You are an expert storyteller. Create a concise meta summary of the given previous summary. \\
\midrule
\textbf{User Prompt} & 
\#\# PREVIOUS\_SUMMARY \newline
[Put your previous summary here] \newline

\#\# Guidelines: \newline
- Write a meta summary of the PREVIOUS\_SUMMARY. \newline
- Focus on key events, character traits, and interactions. \newline

\#\# Output Format: \newline

\#\#\# Meta Summary \newline
[Your meta summary here] \\
\bottomrule
\caption{Iterative Compressor Agent Prompt}
\label{tab:meta_summary_agent}
\end{longtable}
\endgroup

\subsubsection{Iterative Compressor Agent with Few-Shots \texorpdfstring{(\(C'\))}{(C')}}

\begingroup 
\small 
\setlength{\tabcolsep}{4pt}
\small
\begin{longtable}{p{0.15\linewidth}p{0.8\linewidth}}
\toprule
\textit{Single-Turn} & \\
\midrule
\textbf{System Prompt} & 
You are an expert storyteller. Create a concise meta summary of the given previous summary referring the Exmaple Input Output pairs. \\
\midrule
\textbf{User Prompt} & 
\#\# Example Input Output Pairs \newline
\#\#\# Example Previous Summary 01 \newline
[Put your input example] \newline

\#\#\# Exmaple Meta Summary 01 \newline
[Put your output example] \newline

\#\#\# Example Previous Summary 02 \newline
[Put your input example] \newline
 
\#\#\# Exmaple Meta Summary 02 \newline
[Put your output example] \newline

\#\#\# Example Previous Summary 03 \newline
[Put your input example] \newline

\#\#\# Exmaple Meta Summary 03 \newline
[Put your output example] \newline

----------------------------- \newline
Now, Create a concise meta summary of the given previous summary referring the Exmaple Input Output pairs. \newline

\#\# PREVIOUS\_SUMMARY \newline
[Put your previous summary here] \newline

\#\# Guidelines: \newline
- Write a meta summary of the PREVIOUS\_SUMMARY. \newline
- Focus on key events, character traits, and interactions. \newline

\#\# Output Format: \newline

\#\#\# Meta Summary \newline
[Your meta summary here] \\
\bottomrule
\caption{Iterative Compressor Agent Prompt with Few-Shots}
\label{tab:meta_summary_agent_few_shot}
\end{longtable}
\endgroup 

\newpage
\section{Sample Outputs from Each LLM Agent Stage on the MENSA Dataset}
\label{appendix:intermediate_samples}

\subsection{\texorpdfstring{Input Chunk of Narrative (\(N\))}{Input Chunk of Narrative}}

EXT. DEEP SPACE A dark screen is lit up by twinkling stars. SON Baba? FATHER Yes, my son? SON Tell me a story. FATHER Which one? SON The story of home. A meteorite drifts into frame, heading towards tiny Earth off in the distance. FATHER Millions of years ago, a meteorite made of vibranium, the strongest substance in the universe struck the continent of Africa affecting the plant life around it. The meteorite hits Africa and we see plant life and animals affected by vibranium. FATHER -LRB- CONT'D -RRB- And when the time of man came, five tribes settled on it and called it Wakanda. The tribes lived in constant war with each other until a warrior shaman received a vision from the Panther goddess Bast who led him to the Heart Shaped Herb, a plant that granted him super human strength, speed, and instincts. A visual representation of the five tribes emerges as hands from the sand animation, and we see them unite, and then break apart as conflict arises. Bashenga rises above the conflict and eats the Heart Shaped Herb, proceeding to unite the tribes. FATHER -LRB- CONT'D -RRB- The warrior became King and the first Black Panther, the protector of Wakanda. -LRB- MORE -RRB- FATHER -LRB- CONT'D -RRB- Four tribes agreed to live under the King's rule, but the Jabari tribe isolated themselves in the mountains. We see the Jabari striding off towards the isolated mountain region. FATHER -LRB- CONT'D -RRB- The Wakandans used vibranium to develop technology more advanced than any other nation, but as Wakanda thrived the world around it descended further into chaos. We see images of war and slavery just outside Wakanda' s secretive border. FATHER -LRB- CONT'D -RRB- To keep vibranium safe, the Wakandans vowed to hide in plain sight, keeping the truth of their power from the outside world. We see the protective barrier rise around the Wakandan city, as we pull back on the Earth as it spins, now zooming in on another part of the world. . SON And we still hide Baba? FATHER Yes. SON Why?\newline \newline EXT. LAKE MERRITT APARTMENT COMPLEX, OAKLAND - NIGHT - 1992 Towering apartment buildings loom over the horizon. Kids play pickup basketball on a milk carton hoop when mysterious lights approach from the sky.\newline \newline INT. N'JOBU'S APARTMENT/HALLWAY - NIGHT An African man, N'JOBU ( 30s ) sorts through firearms and goes over maps with another man, JAMES ( 20s, African American ). Live news footage of the 1992 Los Angeles Riots plays on the TV. N'JOBU Hey look, if we get in and out quick wo n't be any worries. You in the van come in through from the west, come around the corner, land right here. Me and the twins pulling up right here, we leavin' this car behind, ok? We corning... Suddenly, N'Jobu hears something that we do n't. .. a familiar sound. N'JOBU -LRB- CONT'D -RRB- Hide the straps. James quickly moves into action, hiding the guns in the walls as N'Jobu peers out a window, then clears the maps from the table and unrolls a tapestry on the wall, hiding a bulletin board. James cocks a handgun. JAMES Is it the Feds? N'JOBU No. A KNOCK rattles the door. James walks over and looks out of a peep hole, then turns back with a confused expression. JAMES It's two Grace Jones looking chicks.. they're holding spears... N'JOBU Open it. JAMES You serious? N'JOBU They wo n't knock again. James opens the door and TWO DORA MILAJE ( members of the All Female Wakandan Special Forces ) slowly enter carrying LARGE SPEARS. They eye N' Jobu. DORA MILAJE 1 -LRB- SUBTITLE -RRB- -LRB- in Xhosa -RRB- Who are you? N'JOBU Prince N'Jobu, son of Azzuri. DORA MILAJE 1 -LRB- in Xhosa -RRB- Prove to me you are one of us. N'Jobu grabs his bottom lip and flips it, revealing a vibranium threaded TATTOO glowing blue with Wakandan script. The Dora SWIFTLY SLAM their spears down in unison, making the sound of an EMP. All electronics in the room instantly DIE OUT.\newline \newline INT. N'JOBU'S APARTMENT - NIGHT The room is pitch black, then. . . BOOM. The Dora hit the floor with their spears again and the lights come back ON, but now YOUNG T'CHAKA ( dressed in t he ceremonial garb of the Black Panther ) stands before them. N'Jobu kneels at the sight of him. He swats James, urging him to do the same. N'JOBU -LRB- in Xhosa -RRB- My King... Young T'Chaka looks at James, who gapes in awe. YOUNG T'CHAKA Leave us. N'JOBU This is James. I trust him with my life. He stays, with your permission, King T'Chaka. Young T'Chaka looks at him for a long beat. YOUNG T'CHAKA As you wish. At ease.. Young T'CHAKA removes his mask, hands it off to the Dora and the women step out into the hall, closing the door behind them. YOUNG T'CHAKA -LRB- CONT'D -RRB- -LRB- in Xhosa -RRB- Come baby brother. -LRB- in English -RRB- Let me see how you're holding up. N'Jobu stands. T'Chaka walks over to him and they embrace. YOUNG T'CHAKA -LRB- CONT' D -RRB- You look strong. N'JOBU Glory to Bast. I am in good health. How is home? Young T'Chaka's mood darkens. YOUNG T'CHAKA Not so good, baby brother. There has been an attack. Then, Young T'Chaka activates a simple looking beaded bracelet, KIMOYO BEADS, that project a hologram image of ULYSSES KLAUE. YOUNG T'CHAKA -LRB- CONT'D -RRB- This man, Ulysses Klaue, stole a quarter ton of vibraniurn from us and triggered a bomb at the border to escape. Many lives were lost. He knew where we hid the vibranium, and how to strike. N'Jobu takes the news in. YOUNG T'CHAKA -LRB- CONT'D -RRB- He had someone on the inside. Young T'Chaka waits for N'Jobu to come clean of the crime. He does n't. N'JOBU Why are you here? YOUNG T'CHAKA Because I want you look me in the eyes and tell me why you betrayed Wakanda. N'JOBU I did no such thing. Young T'Chaka snaps a look to James. YOUNG T'CHAKA -LRB- SUBTITLE -RRB- -LRB- in Xhosa -RRB- Tell him who you are. JAMES Zuri, Son of Badu. N'JOBU What? N'Jobu's face goes ashen as James is revealed to be YOUNG ZURI, a Wakandan spy. Zuri reveals his vibranium lip tattoo. N'Jobu grabs him. N'JOBU -LRB- CONT'D -RRB- James, James you lied to me? I invite you into my home and you were Wakandan this whole time? JAMES -LRB- ZURI -RRB- You betrayed Wakanda! N'JOBU How could you lie to me like - YOUNG T'CHAKA Stand down. Did you think that you were the only spy we sent here? Zuri walks to the wall and removes a duffle bag containing VIBRANIUM CANISTERS, glowing blue. He shows one to Young T'Chaka. YOUNG T'CHAKA -LRB- CONT'D -RRB- Prince N'Jobu, you will return home at once, where you will face the council and inform them of your crimes.\newline \newline EXT. LAKE MERRITT, OAKLAND - NIGHT A kid catches the basketball, gazing up as an AIRCRAFT WITH STRANGE LIGHTS rises into the sky and speeds off. The ball drops.\newline \newline INT. COMMAND CABIN, ROYAL TALON FIGHTER - NIGHT - PRESENT DAY T'Challa, dressed in his Black Panther armor, sits at the console in the back of the ROYAL TALON FIGHTER. A BBC news report plays on the screen. BBC ANCHOR The tiny nation of Wakanda is mourning the death of its monarch, King T'Chaka. The beloved ruler was one of many confirmed dead after a terrorist attack at the United Nations a week ago. The Suspect has since been apprehended. -LRB- MORE -RRB- BBC ANCHOR -LRB- CONT'D -RRB- Though it remains one of the poorest countries in the world, fortified by mountain ranges and an impenetrable rain forest, Wakanda does not engage in international trade or accept aid. The succession of the throne is expected to fall to the oldest of the King's two children, Prince T'Challa. Piloting the craft while seated in lotus position is OKOYE ( 30s, Head of the Dora Milaje. ) OKOYE My Prince, coming up on them now.\newline \newline EXT. NIGHT SKY, CHIBOCK - NIGHT An overhead view of a SIX CAR MILITARY CONVOY, moving like a sinister serpent through the surrounding wilderness. We pull back into the sky revealing the tip of a futuristic looking aircraft. This is the ROYAL TALON FIGHTER ( RTF ).\newline \newline INT. COMMAND CABIN, ROYAL TALON FIGHTER - NIGHT T'Challa stands, and moves to a sand model of the convoy down below. Okoye gets up out of her chair and grabs her SPEAR off the wall. She moves towards the back of the craft. T'CHALLA No need, Okoye. I can handle this alone. Okoye pauses, then returns her spear to the wall. T'Challa closes the model and steps to a marked circle in the floor. T'CHALLA -LRB- CONT' D -RRB- I will get Nakia out as quickly as possible. Okoye places six KIMOYO BEADS into T'Challa's gloved hands. OKOYE Just do n't freeze when you see her. T'CHALLA What are you talking about? I never freeze. T'Challa slips on his helmet, revealing himself to be Black Panther and folds his arms over his chest. Okoye opens her closed fist, dropping Panther out of the aircraft.

\subsection{\texorpdfstring{Ouput Chunk Sample from Preprocessor Agent (\(P\))}{Ouput Chunk Sample from Preprocessor Agent}}

EXT. DEEP SPACE\newline \newline A dark screen illuminated by twinkling stars. A young voice, filled with curiosity, broke the silence. "Baba?" The father responded warmly, "Yes, my son?" The boy continued, "Tell me a story." The father asked, "Which one?" The boy's voice lit up with excitement, "The story of home." As he spoke, a meteorite drifted into view, heading towards the distant Earth.\newline \newline The father began his tale, his voice deep and resonant. "Millions of years ago, a meteorite made of vibranium, the strongest substance in the universe, struck the continent of Africa. It affected the plant life around it." The meteorite hit Africa, and the screen showed plant life and animals transformed by vibranium.\newline \newline "And when the time of man came," the father continued, "five tribes settled on it and called it Wakanda. The tribes lived in constant war with each other until a warrior shaman received a vision from the Panther goddess Bast. She led him to the Heart-Shaped Herb, a plant that granted him superhuman strength, speed, and instincts." A visual representation of the five tribes emerged as hands from the sand animation, uniting and then breaking apart as conflict arose. Bashenga, the warrior, rose above the conflict, consumed the Heart-Shaped Herb, and united the tribes.\newline \newline "The warrior became King and the first Black Panther, the protector of Wakanda," the father said, his voice filled with pride. "Four tribes agreed to live under the King's rule, but the Jabari tribe isolated themselves in the mountains." The Jabari were shown striding off towards the isolated mountain region.\newline \newline "The Wakandans used vibranium to develop technology more advanced than any other nation," the father explained. "But as Wakanda thrived, the world around it descended further into chaos." Images of war and slavery flashed just outside Wakanda's secretive border.\newline \newline "To keep vibranium safe, the Wakandans vowed to hide in plain sight, keeping the truth of their power from the outside world." The protective barrier rose around the Wakandan city, and the Earth spun, zooming in on another part of the world.\newline \newline The boy asked, "And we still hide, Baba?" The father replied, "Yes." The boy questioned, "Why?"\newline \newline  EXT. LAKE MERRITT APARTMENT COMPLEX, OAKLAND - NIGHT - 1992\newline \newline Towering apartment buildings loomed over the horizon. Kids played pickup basketball on a milk carton hoop when mysterious lights approached from the sky.\newline \newline  INT. N'JOBU'S APARTMENT/HALLWAY - NIGHT\newline \newline An African man, N'Jobu, in his 30s, sorted through firearms and went over maps with another man, James, a 20-year-old African American. Live news footage of the 1992 Los Angeles Riots played on the TV. N'Jobu explained the plan, "If we get in and out quick, there won't be any worries. You in the van come in from the west, come around the corner, land right here. Me and the twins pulling up right here, we're leaving this car behind, okay?"\newline \newline Suddenly, N'Jobu heard something familiar. "Hide the straps," he ordered. James quickly moved into action, hiding the guns in the walls as N'Jobu peered out a window, then cleared the maps from the table and unrolled a tapestry on the wall, hiding a bulletin board. James cocked a handgun. "Is it the Feds?" James asked. "No," N'Jobu replied.\newline \newline A knock rattled the door. James looked out of a peephole, then turned back with a confused expression. "It's two Grace Jones-looking chicks... they're holding spears..." N'Jobu ordered, "Open it." James hesitated, "You serious?" N'Jobu insisted, "They won't knock again."\newline \newline James opened the door, and two Dora Milaje, members of the All-Female Wakandan Special Forces, slowly entered carrying large spears. They eyed N'Jobu. "Who are you?" one of them asked in Xhosa. N'Jobu replied, "Prince N'Jobu, son of Azzuri." The Dora demanded, "Prove to me you are one of us." N'Jobu grabbed his bottom lip and flipped it, revealing a vibranium-threaded tattoo glowing blue with Wakandan script. The Dora swiftly slammed their spears down in unison, making the sound of an EMP. All electronics in the room instantly died out.\newline \newline  INT. N'JOBU'S APARTMENT - NIGHT\newline \newline The room was pitch black, then... BOOM. The Dora hit the floor with their spears again, and the lights came back on, revealing Young T'Chaka, dressed in the ceremonial garb of the Black Panther, standing before them. N'Jobu kneeled at the sight of him, urging James to do the same. "My King..." N'Jobu said in Xhosa.\newline \newline Young T'Chaka looked at James, who gaped in awe. "Leave us," T'Chaka ordered. N'Jobu pleaded, "This is James. I trust him with my life. He stays, with your permission, King T'Chaka." Young T'Chaka looked at him for a long beat. "As you wish. At ease..."\newline \newline Young T'Chaka removed his mask, handed it off to the Dora, and the women stepped out into the hall, closing the door behind them. "Come, baby brother," T'Chaka said in Xhosa, switching to English, "Let me see how you're holding up." N'Jobu stood, and T'Chaka walked over to him, embracing him. "You look strong," T'Chaka said. N'Jobu replied, "Glory to Bast. I am in good health. How is home?"\newline \newline Young T'Chaka's mood darkened. "Not so good, baby brother. There has been an attack." He activated a simple-looking beaded bracelet, Kimoyo Beads, that projected a hologram image of Ulysses Klaue. "This man, Ulysses Klaue, stole a quarter ton of vibranium from us and triggered a bomb at the border to escape. Many lives were lost. He knew where we hid the vibranium, and how to strike."\newline \newline N'Jobu took the news in. "Why are you here?" he asked. Young T'Chaka replied, "Because I want you to look me in the eyes and tell me why you betrayed Wakanda." N'Jobu denied, "I did no such thing."\newline \newline Young T'Chaka snapped a look to James. "Tell him who you are," he ordered in Xhosa. James revealed, "Zuri, Son of Badu." N'Jobu's face went ashen as James was revealed to be Young Zuri, a Wakandan spy. Zuri revealed his vibranium lip tattoo. N'Jobu grabbed him, "James, James, you lied to me? I invite you into my home and you were Wakandan this whole time?" Zuri retorted, "You betrayed Wakanda!"\newline \newline "How could you lie to me like—" N'Jobu started, but Young T'Chaka interrupted, "Stand down. Did you think that you were the only spy we sent here?" Zuri walked to the wall and removed a duffle bag containing vibranium canisters, glowing blue. He showed one to Young T'Chaka. "Prince N'Jobu, you will return home at once, where you will face the council and inform them of your crimes."\newline \newline  EXT. LAKE MERRITT, OAKLAND - NIGHT\newline \newline A kid caught the basketball, gazing up as an aircraft with strange lights rose into the sky and sped off. The ball dropped.\newline \newline  INT. COMMAND CABIN, ROYAL TALON FIGHTER - NIGHT - PRESENT DAY\newline \newline T'Challa, dressed in his Black Panther armor, sat at the console in the back of the Royal Talon Fighter. A BBC news report played on the screen. "The tiny nation of Wakanda is mourning the death of its monarch, King T'Chaka. The beloved ruler was one of many confirmed dead after a terrorist attack at the United Nations a week ago. The suspect has since been apprehended."\newline \newline The report continued, "Though it remains one of the poorest countries in the world, fortified by mountain ranges and an impenetrable rain forest, Wakanda does not engage in international trade or accept aid. The succession of the throne is expected to fall to the oldest of the King's two children, Prince T'Challa."\newline \newline Piloting the craft while seated in lotus position was Okoye, the Head of the Dora Milaje. "My Prince, coming up on them now," she said.\newline \newline  EXT. NIGHT SKY, CHIBOCK - NIGHT\newline \newline An overhead view of a six-car military convoy moved like a sinister serpent through the surrounding wilderness. The scene pulled back into the sky, revealing the tip of a futuristic-looking aircraft—the Royal Talon Fighter.\newline \newline  INT. COMMAND CABIN, ROYAL TALON FIGHTER - NIGHT\newline \newline T'Challa stood and moved to a sand model of the convoy down below. Okoye got up out of her chair and grabbed her spear off the wall, moving towards the back of the craft. "No need, Okoye. I can handle this alone," T'Challa said. Okoye paused, then returned her spear to the wall. T'Challa closed the model and stepped to a marked circle in the floor. "I will get Nakia out as quickly as possible," he said.\newline \newline Okoye placed six Kimoyo Beads into T'Challa's gloved hands. "Just don't freeze when you see her," she said. T'Challa questioned, "What are you talking about? I never freeze." T'Challa slipped on his helmet, revealing himself to be Black Panther, and folded his arms over his chest. Okoye opened her closed fist, dropping Panther out of the aircraft.

\subsection{\texorpdfstring{Ouput Chunk Sample from Narrative Summarizer Agent (\(S\))}{Ouput Chunk Sample from Narrative Summarizer Agent}}

In deep space, a father tells his son the story of Wakanda, a hidden African nation with advanced technology powered by vibranium, a meteorite that struck the continent millions of years ago. The first Black Panther, a warrior king, united the tribes of Wakanda, except for the isolated Jabari. Wakanda thrived while hiding its power from the chaotic world. The son questions why they still hide.\newline \newline The scene shifts to Oakland, 1992. Prince N'Jobu, a Wakandan spy, plans a heist with James, unaware that James is Zuri, a fellow Wakandan spy. King T'Chaka confronts N'Jobu about a recent attack on Wakanda, accusing him of betrayal. N'Jobu is ordered to return to Wakanda to face the council.\newline \newline In the present day, Prince T'Challa, the new Black Panther, watches a news report about his father's death. He prepares to rescue Nakia, a Wakandan spy, from a militant convoy. Dropping from his aircraft, he uses advanced technology to halt the convoy and begins his rescue mission. Nakia, hidden among the captives, prepares to make her move.

\subsection{\texorpdfstring{Ouput Chunk Sample from First Compressor Agent (\(C_{1}\))}{Ouput Chunk Sample from Compressor Agent}}

In a blend of ancient lore and modern intrigue, a father recounts the origins of Wakanda, a technologically advanced African nation hidden from the world, powered by vibranium. The narrative shifts to 1992 Oakland, where Prince N'Jobu, a Wakandan spy, is confronted by King T'Chaka for betrayal. In the present, Prince T'Challa, the new Black Panther, prepares to rescue Nakia, a Wakandan spy, from a militant convoy, showcasing Wakanda's advanced technology and his own heroism.

\newpage
\twocolumn
\section{\texorpdfstring{Impact of Chunk Size (\(\delta\)) on Summarization}{Analyzing Chunk Size}}
\label{appendix:impact_of_chunk_size}

We analyze the effect of chunk size (\(\delta\)) on compression ratio and summarization quality in LLM-based long-form summarization. \ours~processes all stages of the LLM agent with chunks of input text, then concatenates the resulting outputs to allow for more detailed context analysis by the LLM (see Figure~\ref{fig:overview}). If the compression ratio can be controlled by the chunk size, we can effectively manage the compression level of the final summarization. To demonstrate this relationship, we performed experiments on BookSum, an extensive dataset containing extremely long narratives (>100K). Intermediate outputs are generated using a partial pipeline of \ours{} (consisting of the Preprocessor (\(P\)) and Narrative Summarizer (\(S\))), then these outputs are processed through the Compressor (\(C\)) with varying chunk sizes to analyze the resulting word counts.


\begin{table}[H]
\centering
\small
\begin{tabular}{l|c|c}
\toprule
\textbf{Output of Agent} & \textbf{Word Count} & \textbf{Compression Ratio} \\
\midrule
\(P+S\) & 9675.41 & - \\
\(C_{1} (\delta=500)\) & 4069.24 & 57.94\% \\
\(C_{1} (\delta=1000)\) & 2958.18 & 69.43\% \\
\(C_{1} (\delta=2000)\) & 1967.82 & 79.67\% \\
\(C_{1} (\delta=3000)\) & 1558.41 & 83.88\% \\
\(C_{1} (\delta=4000)\) & 1105.76 & 88.57\% \\
\(C_{1} (\delta=5000)\) & 908.88 & 90.59\% \\
\bottomrule
\end{tabular}
\caption{Analysis of chunk size impact on compression ratio in BookSum. \(P+S\) represents the intermediate output from the Preprocessor (\(P\)) and Narrative Summarizer (\(S\)), while \(C_{1} (\delta=k)\) shows the first-stage compression results with chunk size \(k\). Compression ratio is calculated as \(1 - (C_{1}/P+S)\).}
\label{tab:combined_chunk_size_analysis}
\end{table}

We observe a strong inverse relationship between the chunk size and compression ratio during LLM-based summarization. As shown in Table \ref{tab:combined_chunk_size_analysis}, smaller chunks lead to lower compression ratios, for instance, \(C_{1} (\delta=500)\) compresses input by 57.94\%, whereas \(C_{1} (\delta=5000)\) achieves 90.59\% compression. This suggests that while all chunk sizes remain within the 5K token limit, the model's compression behavior is notably influenced by the granularity of the input narrative segmentation, with smaller chunks preserving more detailed content.

\newpage
\section{\texorpdfstring{Optimal Configuration of Chunk Size (\(\delta\)) }{Optimal Chunk Size}}
\label{appendix:optimal_chunk_size}
We systematically evaluate the impact of chunk size (\(\delta\)) on summarization quality, analyzing performance across four narrative benchmarks. We evaluate each chunk size configuration through a maximum of 10 iterative compression steps without applying a lower bound (\(\theta\)).

\begin{table}[H]
\centering
\scriptsize
\setlength{\tabcolsep}{3pt} 
\renewcommand{\arraystretch}{0.95} 
\begin{tabular}{lcccc}
\toprule
\textbf{Chunk Size (\(\delta\))} & \textbf{BookSum} & \textbf{MovieSum} & \textbf{MENSA} & \textbf{SummScreenFD} \\
\midrule
300 & \textbf{67.16} & \textbf{62.78} & \textbf{65.43} & 58.99 \\
500 & 66.00 & 62.61 & 64.23 & 58.97 \\
1000 & 65.62 & 60.66 & 60.94 & - \\
2000 & 65.32 & 58.07 & - & - \\
4000 & 62.31 & - & - & - \\
8000 & - & - & - & - \\
Maximum(100K) & 50.51 & 55.49 & 55.25 & \textbf{61.59} \\
\bottomrule
\end{tabular}
\caption{Impact of chunk size (\(\delta\)) on summarization quality across benchmarks. Results are reported as BERTScore (F1) after a maximum of 10 iterative compression steps without a lower bound (\(\theta\)). Missing values (-) indicate cases where chunk size exceeds the typical input length for that dataset.}
\label{tab:chunk_size_impact}
\end{table}

As shown in Table~\ref{tab:chunk_size_impact}, BookSum achieves optimal performance (67.16) with a 300-word chunk size, while larger chunks lead to decreased performance, dropping to 62.31 at 4000 words. In the case of MovieSum and MENSA, also 300-word chunk sizes yield peak performances of 62.78 and 65.43. SummScreenFD, designed for shorter content, shows the best score at the Maximum. Three benchmarks demonstrate significantly degraded performance when using maximum chunk size (100K), with scores dropping to 50.51, 55.49, 55.25 for BookSum, MovieSum, MENSA. 

In particular, SummScreenFD presents an inverse relationship with chunk size, achieving optimal performance at maximum context length (61.59 at 100K tokens). This phenomenon aligns with its distinctive characteristics - having the shortest average output length (151 tokens, see Table~\ref{tab:dataset_stats}) among all benchmarks, it benefits from enhanced abstractiveness enabled by larger chunk sizes. This case exemplifies the versatility of \ours{}'s chunk size modulation mechanism, demonstrating its capacity to adapt effectively even to extreme outliers in target summary lengths. Based on the results, we configure \(\delta\)s of 300, 300, 300, and Maximum words for BookSum, MovieSum, MENSA, and SummScreenFD respectively in \ours{}.

\section{\texorpdfstring{{Optimal Configuration of Lower Bound (\(\theta\))}}{Optimal Lower Bound}}
\label{appendix:optimal_lower_bound}


\ours{} achieves superior length control capabilities while maintaining a high BERTScore (F1) compared to Zero-Shot LLM approaches, achieved through lower bound word count constraints in the Compressor (\(C\)) stage (see Section~\ref{subsec:length_control_with_quality_preservation}). We categorize the benchmarks into three distinct length categories: Long-Form (1000-1500 words), Mid-Form (500-1000 words), and Short-Form (0-500 words) following its output average length (see Table~\ref{tab:dataset_stats}). Through systematic experimentation with lower-bound variations at 100-word intervals, we identify optimal configurations for each category. 

\begin{table}[H]
\centering
\scriptsize
\setlength{\tabcolsep}{4pt} 
\renewcommand{\arraystretch}{0.95} 
\begin{tabular}{llccccc}
\toprule
 & \textbf{Lower Bound \(\theta\)} & 1000 & 1100 & 1200 & 1300 & 1400 \\
\midrule
\multirow{1}{*}{\textbf{Long}} & BookSum & 70.69 & 70.42 & 70.54 & \textbf{70.70} & 70.54 \\
\bottomrule
\toprule
& \textbf{Lower Bound \(\theta\)} & 500 & 600 & 700 & 800 & 900 \\
\midrule
\multirow{2}{*}{\textbf{Mid}} & MovieSum & 62.80 & 63.16 & 63.34 & 63.51 & \textbf{63.53} \\
& MENSA & 65.43 & 65.43 & 65.70 & 65.72 & \textbf{65.73} \\
\bottomrule
\toprule
& \textbf{Lower Bound \(\theta\)} & 0 & 100 & 200 & 300 & 400 \\
\midrule
\multirow{1}{*}{\textbf{Short}} & SummScreenFD & \textbf{61.59} & 58.47 & 58.47 & 58.47 & 58.47 \\
\bottomrule
\end{tabular}
\caption{
Optimal Lower Bound (\(\theta\)) configurations across summarization length categories. BERTScore (F1) results are reported for different 
\(\theta\) values: Long-Form (BookSum, 1000-1400 words), Mid-Form (MovieSum and MENSA, 500-900 words), and Short-Form (SummScreenFD, 0-400 words). Bold values indicate the best performance for each benchmark.}
\label{tab:optimal_lower_bound}
\end{table}

As shown in Table~\ref{tab:optimal_lower_bound}, BookSum achieves maximum performance (70.70) with a 1300-word lower bound in the long form category. In the Mid-Form category, both MovieSum and MENSA exhibit optimal performance in the upper range, with scores of 63.53 and 65.73, respectively, in a lower bound of 900 words. For Short-Form summarization, SummScreenFD performs best (61.59) with no lower bound constraint. Based on the results, we set \(\theta\)s of \ours{} for BookSum, MovieSum, MENSA and SummScreenFD to 1300, 900, 900, and 0, respectively.
\begin{table*}[t] 
\centering 
\small 
\setlength{\tabcolsep}{4pt} 
\renewcommand{\arraystretch}{0.80} 
\resizebox{\textwidth}{!}{ 
\begin{tabular}{llc|c|c|c}
\toprule
& & \textbf{BookSum} & \textbf{MovieSum} & \textbf{MENSA} & \textbf{SummScreenFD} \\
& & 1 / 2 / L / (G-M) & 1 / 2 / L  &  1 / 2 / L  &  1 / 2 / L  \\
\midrule
\multirow{4}{*}{\textbf{Zero-Shot}} 
& Mistral Large & 19.63 / 2.99 / 12.0 / (7.87) & \underline{39.22} / \underline{10.53} / \textbf{\underline{22.55}} & \underline{37.43} / \underline{10.52} / 21.52 & 29.18 / 7.43 / \textbf{\underline{19.06}} \\
&  &  &  &  & \\ 
& GPT4o & \underline{20.3} / \underline{3.5} / \underline{17.68} & - & 26.6 / 6.7 / \textbf{\underline{22.5}} & - \\
&  &  &  &  & \\ 
\midrule
\multirow{10}{*}{\textbf{Multi-LLM}} 
& HM-SR & - & 31.31 / 8.81 / 18.62 & 34.26 / 9.74 / 13.46 & - \\
& (GPT4o-mini) &  &  &  & \\ 
&  &  &  &  & \\ 
& CoA Haiku & - / - / - / (13.70) & - & - & - \\
& (Claude 3 Haiku) &  &  &  & \\ 
&  &  &  &  & \\ 
& CoA Sonet & - / - / - / (14.96) & - & - & - \\
& (Claude 3 Sonet) &  &  &  & \\ 
&  &  &  &  & \\ 
& CoA Opus & - / - / - / (\underline{17.47}) & - & - & - \\
& (Claude 3 Opus) &  &  &  & \\ 
&  &  &  &  & \\ 
\midrule
\multirow{4}{*}{\textbf{Ours}} 
& \textbf{\ours{}} & - / - / - / (16.46) & - & - & - \\
& (Claude 3 Haiku) &  &  &  & \\ 
&  &  &  &  & \\ 
& \textbf{\ours{}}
    & \textbf{42.51} / \textbf{10.27} / \textbf{23.91} / \textbf{(18.27)} 
    & \textbf{44.91} / \textbf{11.43} / 19.23 
    & \textbf{48.19} / \textbf{12.50} / 20.48 
    & \textbf{30.44} / \textbf{6.40} / 17.95 \\ 
& (Mistral Large) 
    &&&&\\ 
\bottomrule
\end{tabular}
}
\caption{Performance comparison of \ours{} with LLM-based narrative summarization methods using ROUGE (1/2/L) and Geometric Mean (G-M). Bold indicates best performance, and underlined indicates second-best. \ours{} consistently achieves the highest scores in most metrics across datasets.} 
\label{tab:rouge_results_comparison}
\end{table*}

\newpage
\section{Dataset Contamination Analysis}
\label{appendix:dataset_contamination}
To ensure the reliability of our experimental evaluations, we evaluated potential data contamination in Mistral-Large-Instruct-2407 by conducting an n-gram overlap analysis~\cite{radford2019language} between model-generated summaries and reference summaries in all benchmark datasets. Zero-shot summaries are generated for each dataset using controlled parameters (temperature = 0, top-k = 40) and measured n-gram overlap between generated outputs and reference summaries to detect potential training data leakage.

\begin{table}[H]
\centering
\small
\setlength{\tabcolsep}{3pt} 
\begin{tabular}{l|cccc}
\toprule
\textbf{N} & \textbf{BookSum} & \textbf{MovieSum} & \textbf{MENSA} & \textbf{SummScreenFD} \\
\midrule
4 & 2.00\% & 1.82\% & 1.92\% & 0.66\% \\
5 & 0.95\% & 0.74\% & 0.77\% & 0.26\% \\
6 & 0.46\% & 0.33\% & 0.33\% & 0.11\% \\
7 & 0.24\% & 0.17\% & 0.16\% & 0.04\% \\
8 & 0.15\% & 0.08\% & 0.09\% & 0.02\% \\
\bottomrule
\end{tabular}
\caption{N-gram overlap analysis between model-generated outputs and reference summaries across benchmarks.}
\label{tab:ngram_overlap}
\end{table}

As shown in Table~\ref{tab:ngram_overlap}, overlap rates remain below 2\% across all benchmarks, progressively decreasing for longer n-grams. These low overlap rates indicate negligible contamination, confirming that the base model has not memorized reference summaries, ensuring the integrity of our experimental results.
\newpage
\section{ROUGE Scores Comparison of \ours{} with LLM-Based Narrative Summarization Methods}
\label{appendix:rouge_scores}

Table~\ref{tab:rouge_results_comparison} shows that \ours{} consistently outperforms other LLM-based summarization methods on all benchmarks, achieving the highest ROUGE scores on most metrics. The performance improvements of \ours{} compared to the second-best method in each dataset are as follows:

\begin{itemize}
\item \textbf{BookSum}: ROUGE-1 (+22.21), ROUGE-2 (+6.77), ROUGE-L (+6.23), G-M (+0.8)
\item \textbf{MovieSum}: ROUGE-1 (+5.69), ROUGE-2 (+0.90)
\item \textbf{MENSA}: ROUGE-1 (+10.76), ROUGE-2 (+1.98)
\item \textbf{SummScreenFD}: ROUGE-1 (+1.26)
\end{itemize}

Given the observation that ROUGE scores tend to increase with enhanced detail retention~\cite{chen2024universal}, we propose that \ours{}'s superior performance stems from its ability to capture and incorporate more nuanced and comprehensive features compared to previous LLM-based summarization methods. Further qualitative analysis comparing \ours{} and Zero-Shot methods is provided in Appendix~\ref{appendix:samples_comparsion}.
\section{Factuality Performance and Enhancement}
\label{appendix:factuality_performance}

\ours{} exhibits strong inherent factual consistency, further enhanced through iterative refinement. We evaluated the factual accuracy using NarrativeFactScore ~\cite{jeong2025agentasjudgefactualsummarizationlong}, an LLM-based framework that decomposes summaries into atomic facts and verifies them against the source text. The evaluation is carried out on the MENSA dataset, with GPT4o-mini~\cite{gpt4_technical} as a reference model.

\begin{table}[H]
\centering
\small
\begin{tabular}{l|c}
\toprule
\textbf{Method} & \textbf{NarrativeFactScore} \\
\midrule
TextRank & 59.72 \\
LED & 56.48 \\
LongT5 & 73.76 \\
\midrule
Hierarchically Merging & 81.05 \\
+ Agent Refinement, 1st round & 85.94 \\
+ Agent Refinement, 2nd round & 88.94 \\
+ Agent Refinement, 3rd round & 93.31 \\
\midrule
\ours & 90.16 \\
+ Last Step Refinement, 1st round & 94.43 \\
+ Last Step Refinement, 2nd round & 96.16 \\
+ Last Step Refinement, 3rd round & \textbf{96.83} \\
\bottomrule
\end{tabular}
\caption{NarrativeFactScore results for \ours{} on the MENSA dataset. Without refinement, \ours{} scores 90.16, already surpassing HM-SR (88.94) after two refinement rounds~\cite{jeong2025agentasjudgefactualsummarizationlong}. Further refinement increases accuracy to 96.83, reducing hallucinations.}
\vspace{-10pt}
\label{tab:factuality}
\end{table}

\begin{figure*}[t]
    \centering
    \includegraphics[width=1\linewidth]{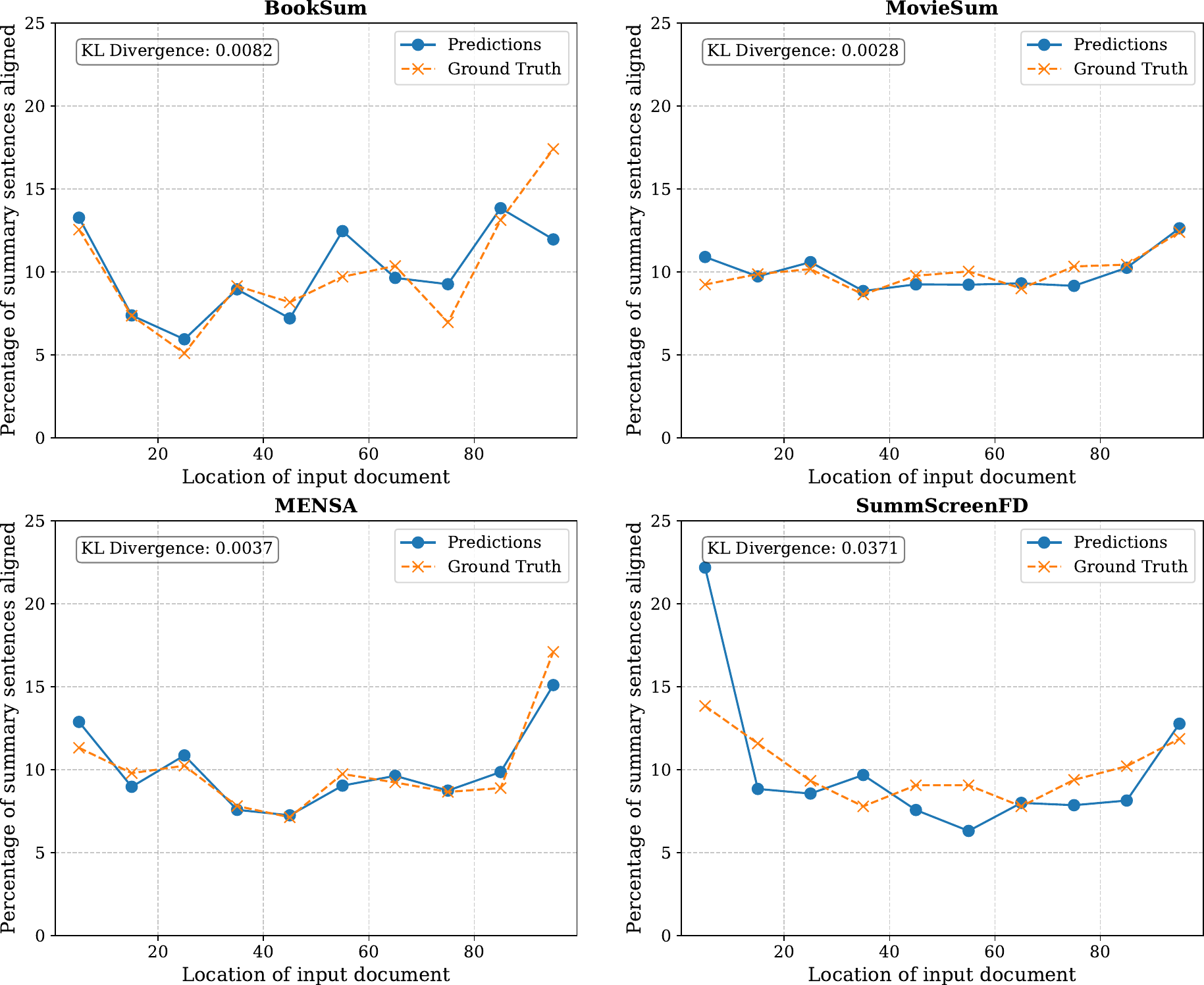}
    \caption{
    Distribution of summary sentence alignments across document locations based on Rouge-L scores. Each subplot shows the percentage of summary sentences mapped to different segments of source documents (divided into ten equal parts). Predictions from \ours{} closely align with human-generated summaries, exhibiting low KL divergence values.}
    \vspace{-10pt}
    \label{fig:utilize_source}
\end{figure*}

Table~\ref{tab:factuality} shows that \ours{} achieves a NarrativeFactScore of 90.16 without explicit refinement, surpassing HM-SR after two rounds of refinement (88.94). Integrating an additional factuality refinement step further boosts the performance to 96.83, demonstrating its effectiveness in reducing hallucinations while maintaining key information.



\section{Document Utilization Analysis}
\label{appendix:document_utilization}

\ours{} effectively extracts key information without additional salient content classification modules, mitgating the "Lost in the Middle" \cite{liu-etal-2024-lost}. To assess information utilization patterns, we analyze the positional distribution of selected content across source documents. 


Following~\cite{saxena2025endtoendlongdocumentsummarization}, we split each source into ten segments and map summary sentences based on the maximum Rouge-L alignment scores. The Kullback-Leibler (KL) divergence is then computed to quantify deviations between predictions and ground-truth summaries. As shown in Figure~\ref{fig:utilize_source}, \ours{}our closely follows human summarization patterns across all benchmarks, with KL divergence values consistently below 0.0371, indicating high alignment with human content selection. 

\section{Inference Time Complexity}
\label{appendix:time_complexity}
We calculate the inference time complexity in Big O notation with respect to the input size ($n$).

\paragraph{Full-context decoder-only LLMs} Full-context decoder-only LLMs, such as those used in zero-shot setups, exhibit the following time complexity when processing inputs of length $n$ ~\cite{zhang2024chain}:

\begin{equation}
T_{\text{total}}^{\text{ZeroShot}} = \mathcal{O}(n^2) + \mathcal{O}(nr)
\end{equation}

where $r$ denotes the average number of response tokens generated. The quadratic term $\mathcal{O}(n^2)$ dominates for long inputs due to attention scaling.

\paragraph{\ours} \ours's hierarchical pipeline consists of three stages: the Preprocessor ($P$), Summarizer ($S$), and Compressor ($C$), each operating on chunked inputs with controlled compression ratios. Let \(c_i\) denote the input chunk tokens of \ours, \(a_i\) represent the compression ratio of responses at each step, and \(r_i\) be the output of each chunk, expressed as \(r_i = a_ic_i\).

The time complexity of the \textbf{Preprocessor ($P$)} for an input of size $n$, with a chunk size of $c_1$ and a compression ratio of $a_1$, is calculated as follows:
\begin{align}
T_{P-\text{Enc}}(n) &= \mathcal{O}(nc_1) \notag \\
T_{P-\text{Dec}}(n) &= \mathcal{O}(nr_1) \notag \\
&= \mathcal{O}(na_1c_1)
\end{align}

For the \textbf{Summarizer ($S$)}, given an input of size $a_1n$ with chunk size $c_2$ and compression ratio $a_2$, the time complexity is:
\begin{align}
T_{S-\text{Enc}}(a_1n) &= \mathcal{O}(a_1nc_2) \notag \\
T_{S-\text{Dec}}(a_1n) &= \mathcal{O}(nr_2) \notag \\
&= \mathcal{O}(a_1a_2nc_2)
\end{align}

For the \textbf{Compressor ($C$)}, given an input size of $a_1a_2n$, a chunk size of $c_3$, and a compression ratio of $a_3$, the time complexity is:
\begin{align}
T_{C-\text{Enc}}(a_1a_2n) &= \mathcal{O}(a_1a_2nc_3) \notag \\
T_{C-\text{Dec}}(a_1a_2n) &= \mathcal{O}(nr_3) \notag \\
&=\mathcal{O}(a_1a_2a_3nc_3)
\end{align}

Finally, the total time complexity of \ours~ is expressed by combining the encoding and decoding steps from each component (Preprocessor ($P$), Summarizer ($S$), Compressor ($C$)) as follows:

\begin{align}
T_{\text{total}}^{\text{\ours}} &= T_{P-\text{Enc}} + T_{P-\text{Dec}} + T_{S-\text{Enc}} \notag \\
 &\quad + T_{S-\text{Dec}} + T_{C-\text{Enc}} + T_{C-\text{Dec}} \notag \\
&= \mathcal{O}(nc_1) + \mathcal{O}(na_1c_1) + \mathcal{O}(a_1nc_2) \notag \\
&\quad + \mathcal{O}(a_1a_2nc_2) + \mathcal{O}(a_1a_2nc_3) \notag \\ 
&\quad + \mathcal{O}(a_1a_2a_3nc_3) \notag \\
&= \mathcal{O}(n( c_1 + a_1 c_1 + a_1 c_2 + a_1 a_2 c_2 \notag \\
&\qquad + a_1 a_2 c_3 + a_1 a_2 a_3 c_3 )) \notag \\
&= \mathcal{O}(n[c_1(1 + a_1) + a_1c_2(1 + a_2) \notag \\
&\qquad + a_1a_2c_3(1 + a_3)])
\end{align}

Given that chunk sizes are fixed hyperparameters and compression ratios are constants, as supported by our empirical analysis in Appendix~\ref{appendix:impact_of_chunk_size}, we can assume that the chunk sizes $c_1, c_2, c_3$ and compression ratios $a_1, a_2, a_3$ are constants. This simplifies \ours's time complexity to:

\begin{align}
T_{\text{total}}^{\text{\ours}} &= \mathcal{O}(n)
\end{align}

Since this linear time complexity can be achieved even when the additional Compressor ($C$) is used, \ours~ achieves an overall time complexity of $\mathcal{O}(n)$, offering substantial scalability advantages for long-form narrative inputs.

\section{\texorpdfstring{Experts Analysis of Zero-shot, \ours~and \ours\textsubscript{R}}{Experts Analysis}}
\label{appendix:qualitative_analysis}

Through the comments of expert evaluators, we qualitatively analyze the generated summaries from each methodology (Zero-Shot, \ours, and \ours\textsubscript{R}). Three K-Drama experts, who evaluate the preference scores, provide comments on the strengths of each approach across the preference criteria:
\begin{itemize}
\item \textbf{Key Events}:

\textbf{Zero-Shot} includes numerous events with concrete scene context summaries, effectively covering sub-plots and script tone through specific examples (e.g., character conflicts). However, excessive minor events obscure key moments despite overall quantity being a strength.

\textbf{\ours} incorporates all critical events concisely, with well-summarized context and implied character motivations. Strengths include capturing hidden narrative details and maintaining clarity without unnecessary elaboration.

\textbf{\ours\textsubscript{R}} covers 80\% of vital relationship-building events crucial for drama plots and includes core aspects of key scenes. Weaknesses include overly condensed critical moments (e.g., flashbacks) and missing worldview-establishing events, impacting atmospheric depth.

\item \textbf{Flow}:

\textbf{Zero-Shot} focuses on overarching drama context, prioritizing character arcs and narrative progression over episode-specific events. Strengths include structured summaries tied to character motivations and long-term plot trajectory. Weakness: Omits granular episode details, sacrificing immediate context for holistic story understanding.

\textbf{\ours} demonstrates clear cause-and-effect relationships in events and provides concrete scene-by-scene context (e.g., character dynamics). Strengths include explicit character behavior rationale and detailed depictions of relationships or personalities. Weakness: Lacks character background context, making motivation interpretation difficult (e.g., supernatural characters' actions).

\textbf{\ours\textsubscript{R}} uses condensed, interpretive descriptions that obscure explicit logical connections (e.g., unclear character perspectives/emotions). While efficiently linking events to specific characters, it fails to adequately highlight protagonist dynamics and situational causes. Key relationship-building contexts remain underdeveloped despite adequate structural hints.


\item \textbf{Factuality}:

\textbf{Zero-Shot}'s factual accuracy is comparable but compromised by the inclusion of content from other episodes and speculative future predictions. While concise summaries of core actions reduce interpretive bias, non-factual forecasts lower overall precision.  

\textbf{\ours} demonstrates high factual accuracy with detailed, episode-specific event listings and clear cause-effect relationships. Avoids speculative content seen in other methods, maintaining strict focus on verified episode 1 details (e.g., no extraneous predictions).

\textbf{\ours\textsubscript{R}} delivers factually correct summaries of key events but introduces subjective emotional interpretations. Context gaps and narrative overstatements slightly undermine reliability despite no outright errors.







\item \textbf{Readability}:

\textbf{Zero-Shot} demonstrates consistently high readability through clear, concise sentences focused on protagonist-driven summaries. Avoids awkward phrasing and excessive detail, making it particularly effective for viewers unfamiliar with the original content to quickly understand character dynamics and plot essentials. 

\textbf{\ours} shows lower readability due to scattered presentation of multiple episodes and subplots that obscure overall narrative context. While containing detailed character interactions, this approach makes key information harder to grasp for audiences unfamiliar with the script.  

\textbf{\ours\textsubscript{R}} achieves mixed results — initial paragraphs are clean and structured, but coherence deteriorates in later sections with abstract phrasing and irrelevant/poetic language. While key elements are condensed for brevity, overcompression hampers overall flow and clarity, especially in conveying contextual nuances.

\end{itemize}

Each summarization approach offers unique strengths, highlighting the potential for future work on \ours~to adapt flexibly to diverse human needs across narrative summarization tasks. While Zero-Shot excels in conciseness, \ours~demonstrates strength in providing detailed information and \ours\textsubscript{R} enhances the summarization process by improving narrative readability while maintaining the detailed enhanced features.


\newpage
\onecolumn
\section{Qualitative Comparison of \ours{} and Zero-Shot Summarization Outputs}
\label{appendix:samples_comparsion}

\subsection{BookSum}

\paragraph{Ground Truth}
The Dashwood family is introduced; Mr. and Mrs. Dashwood and their three daughters live at Norland Park, an estate in Sussex. Unfortunately, Mr. Dashwood's wife and daughters are left with very little when he dies and the estate goes to his son, John Dashwood. John and his wife Fanny have a great deal of money, yet refuse to help his half-sisters and their mother. Elinor, one of the Dashwood girls, is entirely sensible and prudent; her sister, Marianne, is very emotional and never moderate. Margaret, the youngest sister, is young and good-natured. Mrs. Dashwood and her daughters stay at Norland for a few months, mostly because of the promising friendship developing between Elinor and Edward Ferrars, Fanny's shy, but very kind, brother. Elinor likes Edward, but is not convinced her feelings are mutual; Fanny is especially displeased by their apparent regard, as Edward's mother wants him to marry very well. A relative of Mrs. Dashwood's, Sir John Middleton, offers them a cottage at Barton Park in Devonshire; the family must accept, and are sad at leaving their home and having to separate Edward and Elinor. They find Barton Cottage and the countryside around it charming, and Sir John Middleton a very kind and obliging host. His wife, Lady Middleton, is cold and passionless; still, they accept frequent invitations to dinners and parties at Barton Park. The Dashwoods meet Mrs. Jennings, Sir John's mother-in-law, a merry, somewhat vulgar older woman, and Colonel Brandon, a gentleman and a bachelor. The Colonel is soon taken with Marianne, but Marianne objects to Mrs. Jennings attempts to get them together, and to the "advanced" age and serious demeanor of the Colonel. Marianne falls and twists her ankle while walking; she is lucky enough to be found and carried home by a dashing man named Willoughby. Marianne and Willoughby have a similar romantic temperament, and Marianne is much pleased to find that Willoughby has a passion for art, poetry, and music. Willoughby and Marianne's attachment develops steadily, though Elinor believes that they should be more restrained in showing their regard publicly. One pleasant day, the Middletons, the Dashwoods, and Willoughby are supposed to go on a picnic with the Colonel, but their plans are ditched when Colonel Brandon is forced to leave because of distressing news. Willoughby becomes an even more attentive guest at the cottage, spending a great deal more time there than Allenham with his aunt. Willoughby openly confesses his affections for Marianne and for all of them, and hopes they will always think of him as fondly as he does of them; this leaves Mrs. Dashwood and Elinor convinced that if Marianne and Willoughby are not engaged, they soon will be. One morning, Mrs. Dashwood, Elinor, and Margaret leave the couple, hoping for a proposal; when they return, they find Marianne crying, and Willoughby saying that he must immediately go to London. Mrs. Dashwood and Elinor are completely unsettled by this hasty departure, and Elinor fears that they might have had a falling-out. Marianne is torn up by Willoughby's departure, and Elinor begins to question whether Willoughby's intentions were honorable. But, whether Willoughby and Marianne are engaged remains a mystery, as Marianne will not speak of it. Edward comes to visit them at Barton, and is welcomed very warmly as their guest. It is soon apparent that Edward is unhappy, and doesn't show as much affection for Elinor; when they spot a ring he is wearing, with a lock of hair suspiciously similar to Elinor's, even Elinor is baffled. Edward finally forces himself to leave, still seeming distressed. Sir John and Mrs. Jennings soon introduce Mrs. Jennings' other daughter, Mrs. Palmer, and her husband to the family. Mrs. Palmer says that people in town believe that Willoughby and Marianne will soon be married, which puzzles Elinor, as she knows of no such arrangements herself. Elinor and Marianne meet the Middletons' new guests, the Miss Steeles, apparently cousins; they find Miss Steele to be nothing remarkable, while Lucy is very pretty but not much better company. However, the Miss Steeles instantly gain Lady Middleton's admiration by paying endless attention to her obnoxious children. Elinor, unfortunately, becomes the preferred companion of Lucy. Lucy inquires of Mrs. Ferrars, which prompts Elinor to ask about her acquaintance with the Ferrars family; Lucy then reveals that she is secretly engaged to Edward. It turns out that Edward and Lucy knew each other while Edward studied with Lucy's uncle, Mr. Pratt, and have been engaged for some years. Although Elinor is first angry about Edward's secrecy, she soon sees that marrying Lucy will be punishment enough, as she is unpolished, manipulative, and jealous of Edward's high regard for Elinor. The Miss Steeles end up staying at Barton Park for two months. Mrs. Jennings invites Marianne and Elinor to spend the winter with her in London. Marianne is determined to go to see Willoughby, and Elinor decides she must go too, because Marianne needs Elinor's polite guidance. They accept the invitation, and leave in January. Once in town, they find Mrs. Jennings' house comfortable, and their company less than ideal; still, they try their best to enjoy it all. Marianne anxiously awaits Willoughby's arrival, while Elinor finds her greatest enjoyment in Colonel Brandon's daily visits. Elinor is much disturbed when Colonel Brandon tells her that the engagement between Marianne and Willoughby is widely known throughout town. At a party, Elinor and Marianne see Willoughby; Marianne approaches him, although he avoids Marianne, and his behavior is insulting. Marianne angrily writes Willoughby, and receives a reply in which he denies having loved Marianne, and says he hopes he didn't lead her on. Marianne is deeply grieved at being deceived and dumped so coldly; Elinor feels only anger at Willoughby's unpardonable behavior. Marianne then reveals that she and Willoughby were never engaged, and Elinor observes that Marianne should have been more prudent in her affections. Apparently, Willoughby is to marry the wealthy Lady Grey due to his constant need for money. Colonel Brandon calls after hearing the news, and offers up his knowledge of Willoughby's character to Elinor. Colonel Brandon was once in love with a ward to his family, Eliza, who became a fallen woman and had an illegitimate daughter. Colonel Brandon placed the daughter, Miss Williams, in care after her mother's death. The Colonel learned on the day of the Delaford picnic that she had become pregnant, and was abandoned by Willoughby. Elinor is shocked, though the Colonel sincerely hopes that this will help Marianne feel better about losing Willoughby, since he was not of solid character. The story convinces Marianne of Willoughby's guilt, though it does not ease her mind. Out of sympathy, Marianne also stops avoiding the Colonel's company and becomes more civil to him. Willoughby is soon married, which Marianne is grieved to hear; then, again unfortunately, the Miss Steeles come to stay with the Middletons. John and Fanny Dashwood arrive, and are introduced to Mrs. Jennings, and to Sir John and Lady Middleton, deeming them worthy company. John reveals to Elinor that Edward is soon to be married to Miss Morton, an orphan with a great deal of money left to her, as per the plans of his mother. At a dinner party given by John and Fanny for their new acquaintance, Mrs. Ferrars is present, along with the entire Barton party. Mrs. Ferrars turns out to be sallow, unpleasant, and uncivil; she slights Elinor, which hurts Marianne deeply, as she is Edward's mother. The Miss Steeles are invited to stay with John and Fanny. But, Mrs. Jennings soon informs them that Miss Steele told Fanny of Lucy and Edward's engagement, and that the Ferrars family threw the Steele girls out in a rage. Marianne is much grieved to hear of the engagement, and cannot believe that Elinor has also kept her knowledge of it a secret for so long. Edward is to be disinherited if he chooses to marry Lucy; unfortunately, Edward is too honorable to reject Lucy, even if he no longer loves her. Financial obstacles to their marriage remain; he must find a position in the church that pays enough to allow them to marry. Much to Elinor's chagrin, the Colonel, although he barely knows Edward, generously offers the small parish at Delaford to him. Elinor is to convey the offer to Edward, though she regrets that it might help the marriage. Edward is surprised at the generous offer, since he hardly knows the Colonel. Edward decides to accept the position; they say goodbye, as Elinor is to leave town soon. Much to Elinor's surprise, Robert Ferrars, Edward's selfish, vain, and rather dim brother, is now to marry Miss Morton; he has also received Edward's inheritance and money, and doesn't care about Edward's grim situation. It is April, and the Dashwood girls, the Palmers, and Mrs. Jennings, and Colonel Brandon set out for Cleveland, the Palmer's estate. Marianne is still feeling grief over Willoughby; she soon becomes ill after her walks in the rain, and gets a serious fever. The Palmers leave with her child; Mrs. Jennings, though, helps Elinor nurse Marianne, and insists that Colonel Brandon stay, since he is anxious about Marianne's health. Colonel Brandon soon sets off to get Mrs. Dashwood from Barton when Marianne's illness worsens. At last, Marianne's state improves, right in time for her mother and the Colonel's arrival; but Willoughby makes an unexpected visit. Elinor is horrified at seeing him; he has come to inquire after Marianne's health and to explain his past actions. Willoughby says he led Marianne on at first out of vanity; he finally began to love her as well, and would have proposed to her, if not for the money. By saying that he also has no regard for his wife, and still loves Marianne, he attempts to gain Elinor's compassion; Elinor's opinion of him is somewhat improved in being assured of his regard for Marianne. Elinor cannot think him a total blackguard since he has been punished for his mistakes, and tells him so; Willoughby leaves with this assurance, lamenting that Marianne is lost to him forever. Mrs. Dashwood finally arrives, and Elinor assures her that Marianne is out of danger; both Mrs. Dashwood and the Colonel are relieved. Mrs. Dashwood tells Elinor that the Colonel had confessed his love for Marianne during the journey from Barton; Mrs. Dashwood wishes the Colonel and Marianne to be married. Elinor wishes the Colonel well in securing Marianne's affections, but is more pessimistic regarding Marianne's ability to accept the Colonel after disliking him for so long. Marianne makes a quick recovery, thanking Colonel Brandon for his help and acting friendly toward him. Marianne finally seems calm and happy as they leave for Barton, which Elinor believes to signal Marianne's recovery from Willoughby. She is also far more mature, keeping herself busy and refusing to let herself languish in her grief. When Marianne decides to talk about Willoughby, Elinor takes the opportunity to tell her what Willoughby had said at Cleveland, and Marianne takes this very well. Marianne also laments her selfishness toward Elinor, and her lack of civility toward most of their acquaintance. Marianne finally says that she could not have been happy with Willoughby, after hearing of his cruelty toward Miss Williams, and no longer regrets him. The family is stunned when one of their servants returns with news that Edward is married to Lucy, as he just saw them in the village. Elinor knows now that Edward is lost to her forever. Mrs. Dashwood sees how upset Elinor is, and realizes that Elinor felt more for Edward than she ever revealed. One afternoon, Elinor is convinced that the Colonel has arrived at the cottage, but is surprised to find that it is Edward instead. Their meeting is awkward at best; he soon informs them that it is his brother who has been married to Lucy, and not him. Elinor immediately runs from the room, crying out of joy; Edward then senses Elinor's regard for him, and proposes to her that afternoon. Elinor accepts and he gains Mrs. Dashwood's consent to the match. Edward admits that any regard he had for Lucy was formed out of idleness and lack of knowledge; he came to regret the engagement soon after it was formed. After leaving London, Edward received a letter from Lucy saying that she had married his brother Robert, and has not seen her since; thus, he was honorably relieved of the engagement. After receiving the letter, he set out for Barton immediately to see Elinor. Edward will still accept the position at Delaford, although he and Elinor again will not have enough money to live on comfortably. The Colonel visits Barton, and he and Edward become good friends. Edward then becomes reconciled with his family, although he does not regain his inheritance from Robert. His mother even gives her consent for his marriage to Elinor, however much she is displeased by it; she gives them ten thousand pounds, the interest of which will allow them to live comfortably. Edward and Elinor are married at Barton that fall. Mrs. Dashwood and her two remaining daughters spend most of their time at Delaford, both to be near Elinor, and out of the hope that Marianne might accept the Colonel. In the two years that have passed, Marianne has become more mature and more grounded; and she does finally change her mind about the Colonel, and accepts his offer of marriage. The Colonel becomes far more cheerful, and soon Marianne grows to love him as much as she ever loved Willoughby. Mrs. Dashwood remains at Barton with Margaret, now fifteen, much to the delight of Sir John, who retains their company. And Elinor and Marianne both live together at Delaford, and remain good friends with each other and each other's husbands.

\paragraph{Zero-Shot: Mistral-Large-Instruct-2407}
The story follows the Dashwood family as they move to Barton Cottage after the death of their father. The eldest daughter, Elinor, is sensible and responsible, while her younger sister, Marianne, is passionate and romantic. They meet Edward Ferrars, a young man who is secretly engaged to Lucy Steele. Elinor and Edward develop feelings for each other, but Edward is bound by his engagement to Lucy. Meanwhile, Marianne falls in love with John Willoughby, a charming and adventurous young man.\\
Marianne and Willoughby's relationship deepens, and they plan to marry, but Willoughby suddenly leaves for London without explanation. Marianne is heartbroken and falls ill. Elinor discovers that Edward is secretly engaged to Lucy and breaks off their relationship. Colonel Brandon, a friend of the family, reveals that Willoughby has been deceitful and has married a wealthy woman for her fortune.\\
Marianne recovers from her illness and decides to visit London with Elinor. They attend a party where they meet Willoughby, who is now married. Marianne is devastated and leaves the party. Elinor comforts her, and they decide to return to Barton. On their journey home, they meet Colonel Brandon, who reveals his love for Marianne. She is initially upset but eventually accepts his affection. The story ends with the sisters returning to Barton, where they will live happily with Colonel Brandon and Elinor.

\paragraph{\ours{}}
The Dashwood family, particularly Mrs. Dashwood and her daughters Elinor and Marianne, face financial hardship and emotional turmoil after being displaced from their home. John Dashwood, influenced by his wife, fails to support them, leading to their move to a cottage in Devonshire. Elinor, the sensible sister, navigates their struggles while Mrs. Dashwood hopes for a marriage between Elinor and Edward Ferrars, despite his uncertain financial status. The family settles into Barton Cottage, where they meet Sir John Middleton, Colonel Brandon, and Mrs. Jennings. Marianne's musical talents impress Colonel Brandon, sparking matchmaking speculations. Marianne becomes enamored with Willoughby, who rescues her during a rainstorm, while Elinor observes Colonel Brandon's interest in Marianne. The story sets the stage for romantic entanglements and the challenges the Dashwood sisters face in their new environment.
In the previous summary, the dynamics between various characters unfold, highlighting their contrasting personalities and romantic entanglements. Willoughby and Marianne's mutual admiration grows, drawing criticism and ridicule from others, while Elinor defends the reserved Colonel Brandon. Elinor finds herself isolated and seeks companionship with Brandon, discussing Marianne's romantic ideals. Marianne receives and declines a horse from Willoughby, and Elinor suspects an engagement between them.

Colonel Brandon's abrupt departure after receiving distressing news sparks speculation about a mysterious woman named Miss Williams. Willoughby's sudden departure for London leaves Marianne distressed, with Mrs. Dashwood attributing his behavior to external disapproval, while Elinor remains skeptical. Marianne isolates herself in grief, but Mrs. Dashwood remains optimistic about Willoughby's return.

The sisters encounter Edward Ferrars, with Marianne expressing nostalgia for Norland and Elinor remaining practical. Marianne is dissatisfied with their social life, but Elinor defends the Middletons. Edward Ferrars visits the Dashwoods, discussing his lack of ambition and debating with Marianne about wealth and happiness. The contrasting personalities of Elinor and Marianne are highlighted, with Marianne critiquing Elinor's reserved nature. Edward feels misunderstood and introspective, leading to concerns about his mood. He admires practical beauty over the picturesque, disappointing Marianne.
Elinor discovers a ring on Edward's finger, which he claims belongs to his sister, but she suspects it's hers. Social invitations hint at a dance, exciting Marianne due to Willoughby's potential presence. Edward's indecisive behavior, attributed to his mother's influence, leads to his departure. Elinor copes with her grief through activity, while Marianne seeks solitude. The Dashwoods interact with various characters, notably the Palmers and the Miss Steeles, during social gatherings. Key events include Sir John mentioning Marianne's walks to Allenham, sparking interest, and Mrs. Palmer's enthusiastic dinner invitation. The narrative focuses on social dynamics, with Mrs. Palmer praising the Dashwoods and hinting at a rumor about Marianne and Willoughby's potential marriage. The Miss Steeles are introduced and win over Lady Middleton with their fashionable appearance and polite manners.
The Dashwood sisters, Elinor and Marianne, interact with the Miss Steeles, finding the elder unremarkable and the younger, Lucy, lively but lacking elegance. The Miss Steeles ingratiate themselves with Lady Middleton through constant praise. Elinor learns that Edward Ferrars, whom she is fond of, is secretly engaged to Lucy Steele, causing her distress but maintaining composure. Despite Marianne's disdain, Elinor finds Lucy's company tolerable and seeks to understand the engagement. Mrs. Jennings and Sir John tease Elinor about her potential romance with Edward.

The sisters are invited to London by Mrs. Jennings, which Elinor initially declines but ultimately accepts to ensure Marianne's good behavior. During their stay, Marianne grows increasingly anxious waiting for Willoughby, leading to emotional distress when he fails to appear. Elinor worries about Marianne's emotional state and Colonel Brandon's growing affection for her. At social events, Willoughby's cold behavior towards Marianne causes her deep turmoil. Elinor tries to console Marianne but struggles with her own emotions and Mrs. Jennings' obliviousness. Marianne receives conflicting letters from Willoughby, leading to emotional breakdowns.
The previous summary details the emotional turmoil and social dynamics surrounding the Dashwood sisters, particularly Marianne and Elinor. Marianne's intense feelings for Willoughby are revealed through letters, highlighting her imprudence and his insensitivity. Devastated by Willoughby's engagement, Marianne decides to leave for home, struggling with her emotions during a dinner. Mrs. Jennings discusses Willoughby's engagement and hopes for Colonel Brandon's future with Marianne. Elinor comforts her sister and learns more about Willoughby's true character from Colonel Brandon, who reveals Willoughby's mistreatment of his ward, Eliza.

Elinor, practical and composed, navigates complex social dynamics, including the arrival of the Miss Steeles, which causes discomfort due to Lucy's presence. John Dashwood, their brother, displays a mix of concern and arrogance, introducing Elinor to Mrs. Jennings and encouraging her to pursue Colonel Brandon. Elinor encounters Edward Ferrars at a dinner where Mrs. Ferrars shows disdain towards her, yet Elinor remains composed. The summary captures the emotional tensions and social interactions among the Dashwood sisters and their acquaintances, with Elinor focusing on helping Marianne recover.
The previous summary delves into the intricate social and emotional struggles of the Dashwood sisters, Elinor and Marianne. Key events include Elinor's encounters with Edward Ferrars and his family, the revelation of Edward's secret engagement to Lucy Steele, and the sisters' interactions with various characters such as Mrs. Jennings and the Middletons. Elinor's composure and resilience are highlighted as she grapples with keeping Lucy's secret, while Marianne's emotional turmoil and eventual admiration for Elinor's strength are emphasized. The summary also underscores the societal dynamics and familial disapproval surrounding Edward's engagement, with characters like Mrs. Ferrars displaying pride and John Dashwood exhibiting materialism. Despite the turmoil, the sisters defend Edward's integrity, showcasing their contrasting personalities and the complexities of their relationships.
Elinor learns of Edward's engagement to Lucy through gossip and a letter, while Mrs. Jennings offers financial support. Marianne longs for the countryside, leading them to accept an invitation to Cleveland. Colonel Brandon offers Edward a living at Delaford, which Edward gratefully accepts. Mrs. Jennings mistakenly believes Elinor and Brandon are engaged, but Elinor corrects her. Edward expresses gratitude to Brandon, and Elinor reflects on the situation. Mrs. Jennings speculates about Lucy and Edward's future, while Mr. Dashwood is surprised by Brandon's generosity. Elinor discusses Brandon's intentions with a skeptical Mr. Dashwood, who defends Mrs. Ferrars. Robert Ferrars mocks Edward's career choice and disapproves of his engagement to Lucy. The sisters leave London, with Marianne emotional and Elinor relieved. At Cleveland, Marianne falls ill, causing Elinor great anxiety. Colonel Brandon and Mrs. Jennings offer support. Eventually, Marianne shows signs of recovery. Elinor awaits her mother and Colonel Brandon, only to find Willoughby waiting instead. Willoughby confesses his initial pursuit of Marianne was driven by vanity and financial motives, though he developed genuine feelings for her. He admits to his cruelty and selfishness, influenced by past mistakes, and still harbors feelings for Marianne despite his engagement.
The narrative follows the Dashwood sisters, Elinor and Marianne, as they navigate romantic entanglements and personal growth. Elinor discovers that Willoughby's wife intercepted Marianne's letter, leading to a harsh response. Willoughby seeks forgiveness through Elinor, while Mrs. Dashwood arrives, relieved to find Marianne alive and learns of Colonel Brandon's affection for her. Marianne recovers, showing resilience and a new perspective on life. Colonel Brandon's love for Marianne grows, and Elinor eventually shares Willoughby's apology with her sister.

Marianne, initially distressed by Willoughby's actions, resolves to improve her temper and move on. Elinor supports her sister while dealing with her own anxieties about Edward Ferrars. A misunderstanding arises when Edward is seen with Lucy Steele, believed to be his wife, but it is later clarified that Lucy is married to Edward's brother Robert. Edward proposes to Elinor, who accepts, bringing joy to the family. Edward confesses his past mistakes, and the family bonds strengthen.

The story concludes with Edward and Elinor marrying and beginning their life together. The sisters' resilience and the strengthening of family bonds are highlighted throughout the narrative. John Dashwood suggests Marianne pursue Colonel Brandon for his wealth and status, while Robert Ferrars and Lucy Steele regain Mrs. Ferrars' favor.
The narrative explores the dynamics of the Dashwood family, focusing on the romantic entanglements and character traits of its members. Lucy's manipulative tactics secure Ferrars' favor, contrasting with Edward and Elinor's genuine affection. Despite underlying jealousies, the family maintains harmony. Edward is content, and Robert shows no regret over his wealth. Elinor remains close to her family, who visit often. Mrs. Dashwood, Edward, and Elinor hope to unite Marianne and Colonel Brandon, which eventually happens, leading to Marianne's devotion and happiness. Willoughby regrets his past actions but finds solace in his hobbies. Mrs. Dashwood stays at the cottage, while Margaret takes on Marianne's social role. Elinor and Marianne maintain a strong, harmonious relationship, living near each other and preserving family ties.

\subsection{MovieSum}

\paragraph{Ground Truth}
Coraline Jones and her family move from Pontiac, Michigan, to Ashland, Oregon's Pink Palace Apartments. As her parents struggle to complete their gardening catalogue, Coraline is often left alone and meets their new neighbors, including Mr. Sergey Alexander Bobinsky, a Russian circus mouse trainer, Misses April Spink and Miriam Forcible, two once-famous and retired actresses, Wyborne "Wybie" Lovat, the talkative grandson of Pink Palace's landlady, and a mysterious black cat. Wybie gives Coraline a button-eyed rag doll he discovered that eerily resembles her. The doll lures Coraline to a small door in the apartment that is bricked up and can only be unlocked by a button-shaped key. That night, a mouse guides Coraline through the door, now a portal to an “Other World” more colorful and cheerful than her real home. Coraline meets her Other Mother and Father, button-eyed doppelgängers of her parents that appear more attentive and caring. After dinner, Coraline goes to sleep in her Other Bedroom, only to awaken in her real bedroom the next morning. After meeting neighbors Mr Bobinsky and Miss Spink and Forcible, Wybie tells her about his grandmother's twin sister who disappeared in the apartment as a child. Undeterred, Coraline visits the Other World the following two nights, meeting the button-eyed Other Mr Bobinsky, the Other Misses Spink and Forcible, and the Other Wybie, who is mute. On her third visit, the black cat follows her and is able to speak in the Other World. The Other Mother invites Coraline to stay in the Other World forever, on the condition she has buttons sewn over her eyes. Horrified, Coraline attempts to flee but fails. After questioning the other father, Coraline has a conversation with the cat as they walk into the empty part of the world the other Mother created. After breaking the handles of the locked door which leads to the room where the portal is, the Other Mother has blocked the portal and transforms into a menacing version of herself and imprisons Coraline behind a mirror. There, Coraline meets the ghosts of the Other Mother's previous child victims, including the sister of Wybie's grandmother. The spirits reveal that the Other Mother, whom they call the “Beldam,” used rag dolls like Coraline's to spy on them, taking advantage of their unhappy lives and luring them into the Other World with happier and joyful lives. After agreeing to let the Beldam sew buttons on their eyes to let them stay, the Beldam locked them in the mirror and "consumed" their lives, leaving their souls trapped. To free their souls, Coraline promises to find the children's real eyes. Coraline is rescued by the Other Wybie and escapes back to the real world. She discovers her parents are missing, and realizes they have been kidnapped by the Beldam. Miss Spink and Miss Forcible give Coraline an adder stone, telling her that it finds lost things. And she returns to the Other World but falls into a trap. The Beldam locks the door to the portal and swallows its key, but Coraline, following the black cat's advice, proposes a game: if Coraline cannot find her parents and the ghosts' eyes, she will allow her to sew buttons over her own eyes, but if she succeeds, the children's souls will be freed and she can have her real parents back. The Beldam reluctantly agrees and disappears after giving her a clue: "In each of three wonders I've made for you, a ghost's eyes is lost in plain sight."Using the adder stone, Coraline finds the children's eyes and discovers that the Other Wybie was destroyed by the Beldam for helping Coraline escape. As Coraline finds the eyes, the Other World gradually disintegrates until only her family's living room is left. Coraline sees the Beldam in her true skeletal, arachnid form, after showing the Beldam the ghost's eyes, she reminds Coraline that she still has to find her real parents. The ghosts warn her that even if Coraline wins the games, the Beldam will never let her go. Coraline, knowing what to do, tricks the Beldam into unlocking the portal. While the Beldam is distracted, the cat finds her parents trapped in a snow globe, Coraline then throws the cat at the Beldam's face, ripping her button eyes out. Blinded, the Beldam furiously listens for Coraline and almost manages to get her. But Coraline, with help of the ghosts, manages to close the door and lock it but severing the Beldam's left hand. Coraline's parents reappear in the real world with no memory of what happened to them. That night, the ghosts appear in a dream to thank Coraline for freeing their souls and also warn her that the Beldam will never stop looking for the key to the door. As Coraline prepares to drop the key down an old well, the severed hand tries to drag her back to the Other World. But Wybie arrives on his bike and manages to grab the hand with a pair of tongs, but it breaks free, causing him to almost fall into the well. As the hand tries to loosen Wybie's grip for survival, Coraline wraps it in her towel, but it breaks free and it is about to attack her when Wybie smashes it with a rock. They throw the remains of the hand, the key, and the rock into the well and seal it shut. The next day, Coraline and her parents, who have finally finished their catalog, host a garden party for their neighbors. Wybie arrives to the party along with his grandmother, Mrs Lovat. Coraline is happy to meet her and greets her kindly. The camera zooms out from the garden revealing its design which resembles Beldam's face. The camera keeps zooming out to the front of the Pink Palace. The Cat is seen lying on a wooden sign when he stands up and walks into the side vanishing mysteriously, which suggests that he went to the "Other World" through another unknown entrance.

\paragraph{Zero-Shot: Mistral-Large-Instruct-2407}
"Coraline" is a dark fantasy film about a young girl named Coraline Jones who moves with her parents to a new house, the Pink Palace. Bored and neglected by her busy parents, Coraline discovers a small door hidden behind wallpaper in her living room. The door leads to a parallel world where everything seems better: her "Other Mother and Other Father are attentive, the food is delicious, and the garden is magical. However, Coraline soon realizes that this world is a trap set by the Other Mother, a sinister entity who wants to keep her forever by sewing buttons into her eyes.
Coraline meets other children who were previously trapped by the Other Mother and learns that she must find their eyes to free them and escape. With the help of a talking black cat, Coraline navigates the dangerous other world, outsmarting the Other Mother and her minions. She eventually finds the eyes and challenges the Other Mother to a game: if Coraline finds her real parents and the ghost children's eyes, the Other Mother must let them all go.
Coraline succeeds in finding her parents and the eyes, but the Other Mother reneges on their deal. Coraline manages to escape back to her own world, locking the door behind her. However, the Other Mother's hand follows her, trying to retrieve the key. With the help of her friend Wybie, Coraline defeats the hand and throws the key down a well, sealing the Other Mother's fate.
In the end, Coraline's parents return, oblivious to her adventures. Coraline hosts a garden party to celebrate, inviting her new friends and neighbors. The film ends with Coraline enjoying her new life, having learned the value of her real family and the dangers of wishing for something better."

\paragraph{\ours{}}
The story begins with a mysterious transformation of a button-eyed doll in an unknown location. The scene then transitions to the Pink Palace in Ashland, Oregon, where Coraline Jones, an adventurous and curious 11-year-old girl with blue hair, moves in with her parents. Coraline explores her new home and meets Wybie Lovat, a 12-year-old boy who initially appears mysterious but becomes her friend. Wybie shows her an old well in the woods. Confined indoors due to rain, Coraline further explores the house and discovers a small, hidden door in the living room. Intrigued, she asks her mother about it, who reluctantly agrees to find the key. Throughout, Coraline is depicted as a curious and adventurous girl eager to explore her new surroundings.
The narrative introduces Wybie, initially mysterious and annoying, who later reveals himself as a friendly neighbor. The story hints at a deeper mystery involving the Pink Palace and a hidden door in the living room. Coraline and her mother, Mel, discover this door, which initially appears to lead to a brick wall. Later, Coraline follows a ghostly mouse through the door into an alternate, vibrant version of her home. Here, she meets her "Other" parents, who have buttons for eyes and offer her a magical experience. Despite enjoying the evening, Coraline returns home. The next day, her poison oak rash is gone, and she shares her dream with her mother, who suggests visiting their neighbors.
Coraline, a curious and adventurous girl, encounters eccentric neighbors who warn her about a mysterious "little door." She dreams of a parallel world with attentive parents and magical experiences. Feeling misunderstood in her real life, Coraline discovers a key to the little door, leading her to a tunnel and the parallel world from her dreams.
Coraline enters an "Other World" where she meets seemingly perfect "Other Parents" who offer her luxuries. However, they demand she sew buttons into her eyes to stay. Horrified, Coraline tries to escape but is trapped. A talking black cat aids her, revealing the world as an incomplete creation of the Other Mother. Coraline decides to fight back, confronting the Other Mother in a transformed living room. The Other Mother tries to entice her, but Coraline defiantly demands to return to her real parents, angering the Other Mother who grows more menacing.
Coraline is trapped in a mirror closet by the Other Mother (Beldam), where she meets ghost children who reveal that Beldam has imprisoned their souls by sewing buttons over their eyes. The ghosts ask Coraline to find their hidden eyes to free them. With the help of the Other Wybie, Coraline escapes through a dark passageway but is pursued by the Tall Other Mother. She returns to her real home, finding it empty and her parents missing. Wybie arrives and reveals that the doll he gave Coraline belongs to his grandmother, whose missing sister the doll resembles. Coraline shares her eerie experiences with Wybie and believes the sister's ghost is trapped behind the little door in her home.
Coraline and Wybie search for a mysterious doll, which Coraline reveals is a spy for the sinister Other Mother. Wybie, overwhelmed, flees, leaving Coraline frustrated. She discovers her parents' car empty and becomes worried about their disappearance. Later, Miss Spink and Miss Forcible offer Coraline an odd candy, but she leaves, distraught. In her parents' bedroom, she creates makeshift versions of them and falls asleep. A black cat leads her to a mirror, where she sees her parents trapped in a snowy landscape. Realizing the Other Mother has taken them, Coraline burns the button-eyed doll in the fireplace.
Coraline returns to the Other World to rescue her parents, equipped with a bag of items including an odd candy. The cat warns her of a trap but advises her to challenge the Other Mother, who has a weakness for games. Upon entering, Coraline hears her mother's voice and runs towards it. She confronts the Other Mother, who reveals her true sinister form as the Tall Other Mother. Coraline challenges her to a game to find her real parents and the eyes of the ghost children. The Tall Other Mother agrees, hiding the eyes in three "wonders." Coraline navigates these dangerous wonders, outsmarting various twisted versions of characters, and retrieves the ghost eyes. The Other World begins to disintegrate as Coraline discovers her parents trapped in a snow globe and escapes with them, locking the Tall Other Mother in her own world.
Coraline's parents return, unaware of their captivity, leaving her relieved but puzzled. She enjoys a playful moment with her parents, discussing an upcoming garden party. Afterward, she finds mysterious gloves and interacts with a black cat, hinting at lingering connections to the Other World. Determined to free the ghost children's souls, she places their tokens under her pillow and dreams of them as golden angels. They warn her of the Beldam's impending threat. Awakening, Coraline decides to hide the key to the Other World, venturing out despite the cat's attempts to stop her.
Coraline faces the Beldam's dismembered hand, which pursues her to retrieve the key. Wybie, a friend, intervenes but ends up falling into a well with the hand. Together, they defeat the hand and discard its pieces. Wybie apologizes for his earlier disbelief and shares a photo of the sweet ghost girl and her sister. The next day, a garden party celebrates their victory. Coraline expresses gratitude to her neighbors and welcomes Wybie and his grandmother. The scene concludes with the black cat sitting on the Pink Palace Apartments sign, looking into the camera before vanishing.

\subsection{MENSA}

\paragraph{Ground Truth}
Thousands of years ago, five African tribes war over a meteorite containing the metal vibranium. One warrior ingests a "heart-shaped herb" affected by the metal and gains superhuman abilities, becoming the first "Black Panther". He unites all but the Jabari Tribe to form the nation of Wakanda. Over centuries, the Wakandans use the vibranium to develop advanced technology and isolate themselves from the world by posing as a Third World country. In 1992, Wakanda king T'Chaka visits his brother N'Jobu, who is working undercover in Oakland, California. T'Chaka accuses N'Jobu of assisting black-market arms dealer Ulysses Klaue with stealing vibranium from Wakanda. N'Jobu's partner reveals he is Zuri, another undercover Wakandan, and confirms T'Chaka's suspicions. In the present day, following T'Chaka's death, his son T'Challa returns to Wakanda to assume the throne. He and Okoye, the leader of the Dora Milaje regiment, extract T'Challa's ex-lover Nakia from an undercover assignment so she can attend his coronation ceremony with his mother Ramonda and younger sister Shuri. At the ceremony, the Jabari Tribe's leader M'Baku challenges T'Challa for the crown in ritual combat. T'Challa defeats M'Baku and persuades him to yield rather than die. When Klaue and his accomplice Erik Stevens steal a Wakandan artifact from a London museum, T'Challa's friend and Okoye's lover W'Kabi urges him to bring Klaue back alive. T'Challa, Okoye, and Nakia travel to Busan, South Korea, where Klaue plans to sell the artifact to CIA agent Everett K. Ross. A firefight erupts, and Klaue attempts to flee but is caught by T'Challa, who reluctantly releases him to Ross' custody. Klaue tells Ross that Wakanda's international image is a front for a technologically advanced civilization. Erik attacks and extracts Klaue as Ross is gravely injured protecting Nakia. Rather than pursue Klaue, T'Challa takes Ross to Wakanda, where their technology can save him. While Shuri heals Ross, T'Challa confronts Zuri about N'Jobu. Zuri explains that N'Jobu planned to share Wakanda's technology with people of African descent around the world to help them conquer their oppressors. As T'Chaka arrested N'Jobu, the latter attacked Zuri and forced T'Chaka to kill him. T'Chaka ordered Zuri to lie that N'Jobu had disappeared and left behind N'Jobu's American son to maintain the lie. This boy grew up to be Stevens, a black ops U.S. Navy SEAL who adopted the name "Killmonger". Meanwhile, Killmonger kills Klaue and takes his body to Wakanda. He is brought before the tribal elders, revealing his identity to be N'Jadaka and stating his claim to the throne. Killmonger challenges T'Challa to ritual combat, where he kills Zuri, defeats T'Challa, and hurls him over a waterfall to his presumed death. Killmonger ingests the heart-shaped herb and orders the rest incinerated, but Nakia extracts one first. Killmonger, supported by W'Kabi and his army, prepares to distribute shipments of Wakandan weapons to operatives around the world. Nakia, Shuri, Ramonda, and Ross flee to the Jabari Tribe for aid. They find a comatose T'Challa, rescued by the Jabari in repayment for sparing M'Baku's life. Healed by Nakia's herb, T'Challa returns to fight Killmonger, who dons his own nanotech suit, similar to T'Challa's. W'Kabi and his army fight Shuri, Nakia, and the Dora Milaje, while Ross remotely pilots a jet and shoots down planes carrying the vibranium weapons. M'Baku and the Jabari arrive to reinforce T'Challa. Confronted by Okoye, W'Kabi and his army stand down. Fighting in Wakanda's vibranium mine, T'Challa disrupts Killmonger's suit and stabs him. Killmonger refuses to be healed, choosing to die a free man rather than be incarcerated; T'Challa takes him to the waterfall where they fought, where Killmonger dies peacefully. T'Challa establishes an outreach center at the building where N'Jobu died, to be run by Nakia and Shuri. In a mid-credits scene, T'Challa appears before the United Nations to reveal Wakanda's true nature to the world. In a post-credits scene, Shuri helps Bucky Barnes with his recovery.

\paragraph{Zero-Shot: Mistral-Large-Instruct-2407}
"Black Panther" follows T'Challa, the new king of Wakanda, a technologically advanced African nation hidden from the world. T'Challa grapples with the legacy of his father, King T'Chaka, and the responsibilities of his new role. The story begins with T'Challa's father telling him the history of Wakanda and the powerful vibranium that shaped their society. T'Challa learns about his uncle N'Jobu, who betrayed Wakanda by helping Ulysses Klaue steal vibranium, leading to his death at the hands of T'Chaka.
Years later, T'Challa becomes king and faces challenges from within and outside Wakanda. He embarks on a mission to capture Klaue, who has resurfaced, but the mission goes awry when Klaue is rescued by Erik "Killmonger" Stevens, an American black-ops soldier who is revealed to be N'Jobu's son. Killmonger seeks revenge for his father's death and aims to use Wakanda's resources to liberate oppressed people worldwide.
Killmonger challenges T'Challa for the throne and defeats him in ritual combat, seemingly killing him. However, T'Challa is rescued and healed by the Jabari tribe. With the help of his sister Shuri, his love interest Nakia, and his loyal general Okoye, T'Challa returns to reclaim his throne. A climactic battle ensues, with T'Challa ultimately defeating Killmonger but choosing to show him mercy. Killmonger, however, takes his own life.
In the aftermath, T'Challa decides to open Wakanda to the world, sharing their advanced technology and resources to help those in need. The film ends with T'Challa addressing the United Nations, declaring Wakanda's intention to be a beacon of hope and unity.
Key characters include T'Challa, the new king of Wakanda; Killmonger, the vengeful son of N'Jobu; Nakia, T'Challa's love interest and a Wakandan spy; Okoye, the fierce general of the Dora Milaje; and Shuri, T'Challa's brilliant sister and the technological genius behind Wakanda's advancements. The story explores themes of legacy, responsibility, and the balance between tradition and progress.

\paragraph{\ours{}}
The narrative interweaves ancient mythology and contemporary drama, tracing the history of Wakanda, a secretive, technologically advanced African nation fueled by vibranium. In 1992, King T'Chaka confronts Prince N'Jobu for treason in Oakland. In the present, Prince T'Challa, the new Black Panther, embarks on a mission to rescue Nakia, demonstrating Wakanda's advanced tech and his heroism. T'Challa's rescue mission highlights his compassion and combat prowess. Concurrently, Erik Killmonger executes a lethal heist at the British Museum. During his coronation in Wakanda, T'Challa faces a leadership challenge, showcasing his strength and willingness to listen, ultimately becoming king and undergoing a ritual to restore his powers. T'Challa's character is defined by his agility, empathy, and leadership, setting the stage for political intrigue in Wakanda.
In Wakanda, key characters emerge: Nakia, a skilled fighter and advocate; Okoye, a loyal warrior; Killmonger, a cunning and ruthless figure with Wakandan ties; and M'Baku, a formidable and honorable challenger. T'Challa, the new king, seeks guidance from his father's memory. T'Challa and Nakia debate Wakanda's future, with Nakia advocating for resource sharing and T'Challa being cautious. W'Kabi, T'Challa's friend, prefers a more aggressive global approach. The Tribal Council learns of Ulysses Klaue's plan to sell a stolen Wakandan artifact in South Korea. T'Challa, driven to capture Klaue, prepares with advanced technology from his sister Shuri, including a new Black Panther suit. In Busan, T'Challa, Okoye, and Nakia infiltrate a casino to intercept Klaue, encountering CIA agent Everett Ross, who is also after Klaue.
In a dynamic sequence, a fight erupts upon Okoye's discovery, leading to a high-speed car chase through the city. T'Challa, as Black Panther, joins the pursuit using a remotely driven car operated by Shuri. The chase splits, with Nakia and Okoye pursuing one group while Black Panther chases Klaue. T'Challa and Shuri demonstrate their combat and driving skills, with T'Challa utilizing his suit's kinetic energy to disable multiple SUVs. Nakia and Okoye, protected by their vibranium car, remain unfazed by gunfire, with Okoye impressively disabling an SUV with her spear. Despite their relentless efforts and showcasing their unique abilities, Klaue, equipped with a sonic disrupter, gains the upper hand, causing chaos and dismantling Nakia and Okoye's car. The chase underscores the characters' determination and skills but concludes with Klaue's dangerous tactics prevailing.
In a dynamic sequence, Okoye showcases her agility while Ross joins the pursuit of Klaue. T'Challa and Shuri collaborate to track Klaue's SUV through Centum City Bridge, with Shuri using advanced tech to keep pace. The high-speed chase, involving the Black Panther in a Lexus LC, culminates in the Haeundae District where Klaue's sonic blast destroys the car, but Shuri survives and T'Challa captures Klaue. At a CIA black site, Klaue reveals Wakanda's secrets to Agent Ross. Killmonger and his team attack the site, freeing Klaue but ultimately killing him and his own team.
Ross is critically injured and taken to Wakanda for healing, causing tension with Okoye. Killmonger's true intentions and lineage are revealed, as he is N'Jobu's son and seeks the throne. T'Challa learns about his uncle's betrayal and his father's past actions, causing him to question his legacy. Killmonger arrives in Wakanda with Klaw's corpse, while Ross awakens in Shuri's lab and discovers Wakanda's advanced technology. Shuri and Ross uncover Killmonger's identity and his claim to the throne. T'Challa and Nakia discuss T'Chaka's past, with Nakia encouraging T'Challa to forge his own path as king.
Killmonger, revealed as N'Jadaka, challenges T'Challa for Wakanda's throne, aiming to use its resources to liberate the oppressed. Despite his mother's protests, T'Challa accepts the challenge. During the ritual combat at Warrior Falls, Zuri intervenes but is killed by Killmonger. T'Challa is defeated and presumed dead, and Killmonger is crowned king. Okoye struggles with her loyalty, while Nakia, Ross, Ramonda, and Shuri unite to resist Killmonger's rule. A flashback shows young Killmonger discovering his father's death and Wakandan secrets. Adult Killmonger undergoes a transformative ritual, emerging empowered. He orders the destruction of the Heart Shaped Herbs, symbolizing a break from tradition and the start of his new regime.
Nakia escapes with a crucial herb. Killmonger reveals his global armament plan to create an empire. Okoye and W'Kabi debate Wakanda's future, with Okoye favoring tradition and W'Kabi supporting Killmonger. Nakia, Shuri, Ramonda, and Ross find T'Challa alive but comatose and revive him. T'Challa resolves to reclaim his throne. Killmonger begins distributing vibranium weapons globally. T'Challa confronts Killmonger, sparking a major battle. W'Kabi and the Border Tribe side with Killmonger, while Okoye and the Dora Milaje support T'Challa. Shuri, Nakia, and Ross work to intercept weapon shipments. The battle includes armored rhinos and aerial combat.
In the climactic sequence, T'Challa and his allies face multiple fronts to save Wakanda and the world. T'Challa confronts W'Kabi, while Okoye and the Dora Milaje battle Killmonger. Nakia and Shuri, with Ross's help, commandeer a fighter jet to stop vibranium-laden aircraft. The story culminates in a decisive duel between T'Challa and Killmonger in a vibranium mine. Meanwhile, Shuri, Nakia, Okoye, and Ayo join forces with M'Baku and the Jabari tribe against W'Kabi and the Border Tribe. T'Challa defeats Killmonger, who chooses death over imprisonment. T'Challa then announces Wakanda's global outreach, with Shuri and Nakia taking key roles, showcasing their leadership and expertise.
The script concludes with T'Challa, Nakia, Okoye, and Ayo at the United Nations, where T'Challa delivers an inspiring speech promoting global unity and cooperation. Additionally, Bucky Barnes is depicted in Wakanda, recovering and learning under Shuri's guidance, hinting at his integration into Wakandan society.

\subsection{SummScreenFD}

\paragraph{Ground Truth}
Michael mourns the death of his comic book hero, Captain Astro, while Debbie finds a real life murder victim behind the diner. Emmett helps George to enjoy himself. Melanie \& Lindsay are inspired by the contents of Lindsay's Granny Faye's secret letters.

\paragraph{Zero-Shot: Mistral-Large-Instruct-2407}
The script follows two main storylines involving a group of LGBTQ+ friends and their personal struggles. The first plot revolves around Michael, who is devastated by the death of his childhood hero, Captain Astro, a comic book character. Michael's friends, including Brian, Justin, Emmett, and Ted, try to console him, but Michael remains distraught. His mother, Debbie, criticizes him for caring more about a fictional character than a real person after a young man is found dead outside her diner.
Meanwhile, Lindsay and Melanie are planning their wedding but face financial constraints and disagreements over the menu and Lindsay's desire to wear her grandmother's wedding dress. They discover old letters revealing that Lindsay's grandmother had a lesbian relationship, which brings them closer together.
Emmett introduces his new friend, George, an older man, to the group. George, who has been living a reclusive life, is encouraged by Emmett to attend the opera, where they encounter George's ex-wife, Virginia. Emmett stands up for George, leading to a heated exchange with Virginia.
Brian and Justin engage in a dangerous sexual encounter, highlighting the risks they take in their relationship. Michael, feeling disillusioned, seeks solace on a rooftop overlooking Babylon, a gay club. Ben, Michael's partner, comforts him, and they decide to have their own candlelight vigil for the young man found dead.
The script explores themes of loss, identity, and the complexities of relationships within the LGBTQ+ community. It underscores the importance of support and understanding among friends and partners, even in the face of societal and personal challenges.

\paragraph{\ours{}}
Michael and his friends mourn the death of Captain Astro, a beloved comic book hero, while Debbie discovers a body in the dumpster behind the diner. Lindsay and Melanie argue over wedding plans, leading to a discovery about Lindsay's grandmother's secret past.

\end{document}